\documentclass{article}

\usepackage{PRIMEarxiv}

\usepackage[utf8]{inputenc}
\usepackage[T1]{fontenc}

\usepackage{graphicx}%
\usepackage{multirow}%
\usepackage{amsmath,amssymb,amsfonts}%
\usepackage{booktabs}%
\usepackage{xcolor}%
\usepackage{colortbl}%
\usepackage{array}%
\usepackage{makecell}%
\usepackage{pifont}%
\usepackage{enumitem}%
\usepackage{placeins}%
\usepackage{threeparttable}%

\usepackage[round]{natbib}

\usepackage{url}
\usepackage[hidelinks]{hyperref}

\begin{document}

\title{When Do VLMs Help Arabic Manuscript OCR? A Cross-Dataset Study\thanks{This is a preprint of a manuscript currently under peer review. It has not yet been peer-reviewed and may differ substantially from later versions.}}

\author{
  Moshiur Farazi$^{1,*}$ \quad
  Firoj Alam$^{2}$ \quad
  Abderrahmane Maaradji$^{1}$ \quad
  Zakaria Maamar$^{1}$ \\[3pt]
  Hamdy Mubarak$^{2}$ \quad
  Wajdi Zaghouani$^{3}$ \\[6pt]
  \normalfont\small
  $^{1}$University of Doha for Science and Technology, Doha, Qatar \\
  $^{2}$Qatar Computing Research Institute, Hamad Bin Khalifa University, Doha, Qatar \\
  $^{3}$Northwestern University in Qatar, Doha, Qatar \\[3pt]
  $^{*}$Corresponding author: \texttt{moshiur.farazi@udst.edu.qa}
}

\date{}

\maketitle

\begin{abstract}
Vision-language models (VLMs) are increasingly being used for document understanding, yet their role in Arabic and Islamic manuscript recognition remains underexplored. To address such a gap in this paper, we evaluate traditional OCR, general-purpose VLMs, Arabic-specialized VLMs, and OCR-conditioned VLM correction across eight Arabic text datasets spanning historical manuscripts, aged printed books, clean print, multi-domain documents, and handwriting. The results show that no single approach dominates across setups. On line-level historical manuscripts, VLMs are close to Tesseract; on page-level manuscript images, they perform better; and in several settings, an OCR-conditioned corrector improves over both standalone OCR and standalone VLMs. The central finding is an OCR-prior recoverability principle: OCR conditioning helps when the OCR output remains visually and textually recoverable, providing anchors that the VLM can refine against the image. It improves recognition on aged print, clean print, mixed-domain Arabic, and some Naskh manuscripts, but degrades performance when the prior is script-mismatched or systematically misleading, as in Maghribi manuscripts and realistic student handwriting. Additional diagnostics show that Arabic VLM-OCR is sensitive to diacritics, preprocessing, generation budget, and repetition loops. These findings support an adaptive OCR-VLM workflow that routes pages according to script, OCR-prior recoverability, length diagnostics, and failure-mode indicators.

\end{abstract}


\section{Introduction}\label{sec_introduction}

Vision-Language Models (VLMs) are changing how document understanding systems are designed. Text recognition is moving from OCR-dependent pipelines toward OCR-free and end-to-end models that can read visually rich pages and reason over their content~\citep{kim2022donut,wei2024gotocr,bai2025qwen25vl}. Recent work on historical documents shows that multimodal LLMs can transcribe archival records competitively, and in some Latin-script settings they can outperform conventional OCR and Handwritten Text Recognition (HTR) systems~\citep{greif2025mllm_historical,semnani2025churro}. The main question is therefore no longer only whether VLMs can read historical documents. It is also when they should be used, whether they should replace or complement OCR, and how their outputs should be evaluated~\citep{greif2025mllm_historical,kanerva2025ocr}. For Islamic manuscripts, these deployment questions remain largely unanswered. Existing Arabic OCR and VLM work focuses mainly on modern Arabic OCR, handwriting, and document understanding, while historical Arabic work has focused mainly on datasets and supervised HTR rather than zero-shot VLM-based manuscript recognition~\citep{bhatia2024qalam,heakl2025kitab,muharaf2024,chan2024hatformer}.

This gap matters for cultural heritage. Islamic manuscripts preserve one of the world's richest textual traditions. They span Qur'anic sciences, hadith, jurisprudence, theology, medicine, astronomy, philosophy, literature, and the transmission of scholarly knowledge across centuries~\citep{gruber2010islamic,gacek2009arabic}. Major libraries and digital initiatives have invested heavily in preservation and access. Large collections are held by the British Library, the Qatar Digital Library, the Biblioth\`{e}que nationale de France, the Bodleian Library, and institutions across the Middle East and North Africa. The Qatar Digital Library alone contains more than two million digitised pages~\citep{qnl2026heritage}. These efforts have made high-resolution manuscript images increasingly available. Yet digitised images alone do not make these collections computationally searchable, linkable, or analysable.

The main bottleneck is the lack of reliable machine-readable structure extracted from the images. Researchers, curators, and collection users still rely heavily on catalogue descriptions and keyword search. Manuscript pages contain much richer evidence, including transcribed text, layout, marginalia, colophons, script styles, dates, places, names, and subject cues~\citep{gacek2009arabic,quiring2013colophon}. Recovering this evidence requires systems that can move from pixels to text and from text to structured knowledge~\citep{abdallah2024survey,borchmann2021measuring}.

\begin{figure}[ht]
\centering
\includegraphics[width=0.8\columnwidth]{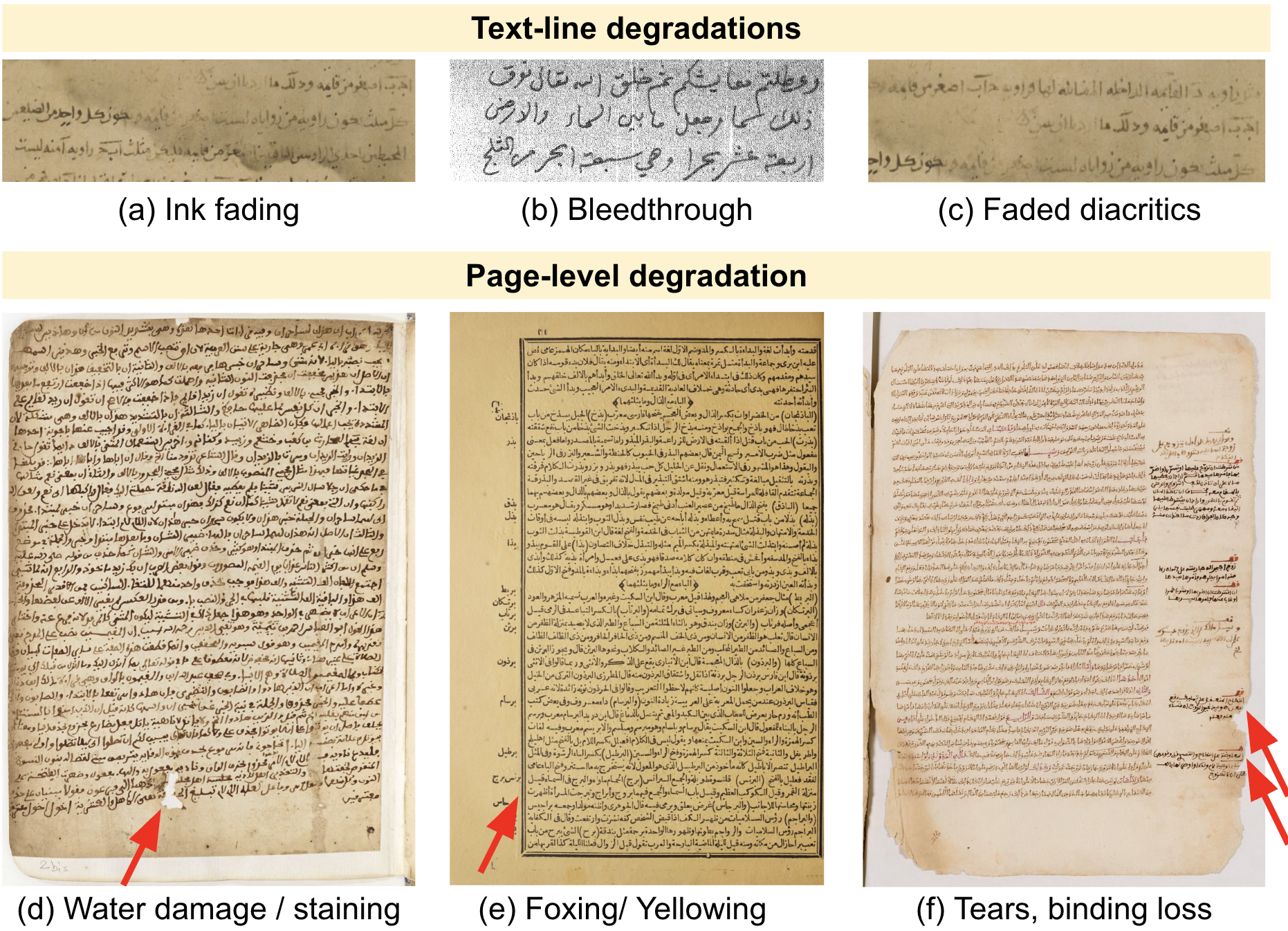}
\caption{Representative data-level degradation modes in Arabic manuscript scans, drawn from samples in our evaluation corpora. Each mode corresponds to a failure pattern that traditional OCR and VLM systems handle differently.
}
\label{fig_data_challenges}
\end{figure}

This transfer from image to machine-readable text is not guaranteed for Islamic manuscripts. Arabic is cursive and context-sensitive. Letter shapes depend on position, and small dots or diacritics can change character identity, pronunciation, or meaning~\citep{faizullah2023survey,bhatia2024qalam}. Manuscript images add further difficulty (Figure~\ref{fig_data_challenges}). They often contain marginal glosses, commentary around a base text, rubrication, colophons, paratextual notes, degraded paper, faded ink, and script traditions such as Naskh and Maghribi~\citep{gacek2009arabic,quiring2013colophon}. These visual and linguistic properties make it unclear whether VLM gains reported for Latin-script historical OCR will transfer to Arabic-script manuscript collections~\citep{greif2025mllm_historical,semnani2025churro}.

The missing piece is not Arabic OCR itself. Recent Arabic-specific systems such as Qalam, QARI-OCR, and Baseer show rapid progress on Arabic OCR, handwriting recognition, and document-to-Markdown conversion~\citep{bhatia2024qalam,qariocr2025,baseer2025}. KITAB-Bench and SARD provide modern multi-domain and synthetic printed Arabic benchmarks~\citep{heakl2025kitab,sard2025}. Historical Arabic work has mainly produced datasets and supervised HTR systems, including Muharaf and HATFormer~\citep{muharaf2024,chan2024hatformer}. What remains unclear is whether current VLMs can serve as standalone recognisers or OCR-conditioned correctors for Islamic manuscript digitisation. It is also unclear which document settings favour each strategy.

This paper addresses that gap through a cross-domain zero-shot evaluation of traditional OCR, general-purpose VLMs, Arabic-specialised VLMs, and OCR-conditioned VLM correction. We evaluate eight Arabic text datasets with more than 7,700 samples. The datasets cover historical manuscripts, aged printed books, modern handwriting, multi-domain Arabic documents, and synthetic clean print. We treat these datasets as distinct document settings rather than as one pooled benchmark, since each setting tests a different deployment condition.

\noindent
\textbf{Research questions.}
We ask five questions.

\begin{itemize}[noitemsep,topsep=0em,leftmargin=1.5em,labelsep=.5em]
\item \textbf{RQ1.} How do zero-shot VLMs compare with traditional OCR and supervised HTR baselines on Arabic manuscript recognition.

\item \textbf{RQ2.} When does OCR-conditioned VLM correction improve recognition, and when does the OCR prior mislead the model.

\item \textbf{RQ3.} Do common image preprocessing operations improve VLM recognition on historical Arabic manuscript lines.

\item \textbf{RQ4.} Which failure modes shape Arabic VLM-OCR, especially for diacritics, hallucination, truncation, and repetition.

\item \textbf{RQ5.} What does a zero-shot metadata extraction pilot reveal about prompt sensitivity and compliance-bias hallucination.

\end{itemize}

\noindent
\textbf{Contributions.}
The paper makes five contributions.

\begin{itemize}[noitemsep,topsep=0em,leftmargin=1.5em,labelsep=.5em]
\item We provide a cross-domain zero-shot benchmark of traditional OCR, general-purpose VLMs, Arabic-specialised VLMs, and OCR-conditioned VLM correction across eight Arabic text datasets.

\item We introduce a hybrid OCR-corrector pipeline in which a VLM receives both the source image and noisy Tesseract output, and we show through ablations that both signals are necessary for its gains.

\item We identify an OCR-prior recoverability principle. OCR conditioning helps when the OCR output remains visually and textually recoverable, but it can hurt when the prior is script-mismatched or systematically misleading.

\item We analyse deployment-critical failure modes, including diacritics sensitivity, preprocessing effects, generation-budget truncation, repetition-loop failures, and metadata compliance bias.

\item We provide an exploratory metadata extraction study showing that null-enforcing prompt calibration can reduce schema-filling behaviour on manuscript pages.

\end{itemize}

\noindent
\textbf{Main findings.}
The experiments answer the research questions directly. For \textbf{RQ1}, zero-shot VLMs are not yet replacements for supervised HTR, but they are competitive with traditional OCR in several manuscript settings. On line-level Muharaf manuscripts, Qwen2.5-VL-7B is close to Tesseract after diacritics normalisation. On page-level Madinah images, VLMs perform better than Tesseract, suggesting that page context can help VLM recognition. For \textbf{RQ2}, OCR-conditioned correction helps only when the OCR prior remains recoverable. On Muharaf, Tesseract$\rightarrow$Qwen reduces CER from 69.2\% to 60.6\%. The corrector also improves SARD, KITAB-Bench, and QNL Books. In contrast, it hurts or gives little benefit on RASAM~2, KHATT, and Students Essays, where the OCR prior is less reliable or script-mismatched.

For \textbf{RQ3}, common OCR preprocessing does not uniformly help VLM recognition. On Muharaf, sharpening is the only tested operation that improves Qwen, reducing CER from 70.4\% to 67.2\%. Binarisation, denoising, contrast enhancement, and combined pipelines degrade performance, suggesting that aggressive preprocessing can damage small Arabic-script cues. For \textbf{RQ4}, Arabic VLM-OCR is shaped by failure modes that CER alone does not fully explain. Diacritics add a large CER penalty for the main VLMs, while Tesseract often avoids this penalty by omitting tashkeel. Page-level generation is also sensitive to decoding settings. A small generation budget causes truncation on SARD, while greedy decoding causes repetition loops on about 28\% of Students Essays pages. For \textbf{RQ5}, the metadata pilot shows that zero-shot structured extraction is highly prompt-sensitive. A permissive prompt makes the VLM fill metadata fields with body text, while a null-enforcing prompt reduces this compliance-bias behaviour. We therefore treat this experiment as a failure-mode study rather than as a full metadata benchmark.

Overall, the findings support an adaptive OCR-VLM workflow. A target collection should first be tested on a small annotated probe sample. The probe should compare traditional OCR, standalone VLMs, and OCR-conditioned correction, and it should inspect CER, length ratio, repetition, script type, and qualitative OCR-prior recoverability. When the OCR prior remains readable and recoverable, the corrector is useful. When the prior is distorted, script-mismatched, or dominated by systematic errors, a standalone VLM is safer.

The remainder of this paper is organised as follows. Section~\ref{sec_background_related} reviews existing approaches to text recognition, layout analysis, and metadata extraction for Arabic manuscripts, and presents a taxonomy of challenges. Section~\ref{sec_datasets} describes the datasets. Section~\ref{sec_experimental_design} describes the experimental design. Section~\ref{sec_results} reports the results. Section~\ref{sec_discussion} discusses deployment implications, limitations, and future directions. Section~\ref{sec_conclusion} concludes the paper.

\section{Background and Related Work}
\label{sec_background_related}

Islamic manuscript recognition sits at the intersection of Arabic OCR, historical handwritten text recognition, document layout analysis, metadata extraction, and vision-language modelling~\citep{Bhatia_2026}.

\subsection{Why Islamic Manuscript Images Are Difficult}
\label{sec_background_challenges}

Islamic manuscripts differ from both modern Arabic documents and many Latin-script historical records. The system input is a digitised page image, but many recognition difficulties come from the manuscript object itself.

Arabic script is cursive and context-sensitive. Most letters connect to neighbouring letters within a word, and the same letter may take isolated, initial, medial, or final forms depending on its position. This makes character segmentation ambiguous and error-prone, which has long been a central challenge in Arabic OCR~\citep{faizullah2023survey,bhatia2024qalam}.

Small visual marks also carry substantial information. Dots distinguish many Arabic letters, and diacritics, or \textit{tashkeel}, can change pronunciation and meaning. In Qur'anic and scholarly manuscripts, diacritics may be part of the text rather than optional decoration. They are also difficult to recognise visually because they are small and are often affected by fading, scanning noise, or paper degradation.

Historical Islamic manuscripts show wide script variation. Common traditions include Naskh, Maghribi, Thuluth, Kufic, Ta'liq, and Nasta'liq. These scripts differ in letter shape, proportion, spacing, and decorative practice~\citep{gacek2009arabic}. A system that works on one script style may not transfer well to another. This is especially important for historical Arabic HTR, where supervised systems can perform well in-domain but often depend on labelled data from the target script tradition~\citep{chan2024hatformer,rasam2024}.

The page layout adds another source of difficulty. Many manuscripts contain marginal glosses, commentary around a base text, rubrication, decorative headings, ownership notes, and colophons. A single page may contain several text zones with different functions and reading orders. Colophons are especially important because they may contain names, dates, places, and copyist information~\citep{quiring2013colophon}. These features make simple line-by-line OCR insufficient for search, indexing, and downstream analysis.

The content also differs from modern web Arabic. Many manuscripts use classical Arabic, specialised terminology, and dense scholarly conventions. They may contain Qur'anic quotations, hadith chains, legal concepts, scientific terminology, names of authorities, and transmission certificates. A useful system must therefore preserve the written text accurately while also supporting later linking, classification, and metadata extraction.

Finally, many manuscript images are physically degraded. Ink fading, bleedthrough, stains, foxing, tears, page curvature, and binding damage are common. These artefacts interact with the visual density of Arabic script. They can hide dots, erase diacritics, or break thin connecting strokes. As a result, a system that works on clean printed Arabic may fail on historical manuscript pages.

\subsection{Arabic OCR and Historical HTR}
\label{sec_background_ocr_htr}

Traditional Arabic OCR systems usually follow a pipeline. They apply preprocessing, layout analysis, character or word recognition, and post-processing. Errors introduced at one stage can affect all later stages~\citep{faizullah2023survey,kasem2023arabic,gupta2007ocr_binarization}. Tesseract remains a widely used open-source OCR system and supports Arabic~\citep{smith2007tesseract}. It is therefore a useful reference point for practical digitisation projects, even when it is not expected to match supervised HTR on difficult manuscripts.

Supervised HTR systems can be much stronger when labelled training data is available. CNN, RNN, CTC, and transformer-based systems have been widely used for offline handwriting recognition~\citep{graves2008offline}. For historical Arabic, HATFormer reports strong supervised performance on Muharaf~\citep{chan2024hatformer}. RASAM~2 also supports supervised recognition of Maghribi manuscripts~\citep{rasam2024}. These results show the value of domain-specific training.

The main limitation is data availability. Most Islamic manuscript collections do not have enough expert transcriptions to train a collection-specific HTR model. Script traditions, page layouts, and physical degradation also vary across collections. This makes supervised training expensive and difficult to scale. It also motivates zero-shot and weakly supervised alternatives that can be deployed before large annotation efforts are possible.

\subsection{VLMs for Document and Historical OCR}
\label{sec_background_vlms}

Recent VLMs have changed how document understanding is framed. Earlier systems often depended on OCR before downstream processing. Newer systems treat recognition and understanding as a joint vision-language task. Donut introduced OCR-free document understanding~\citep{kim2022donut}. GOT-OCR frames OCR as a unified end-to-end task~\citep{wei2024gotocr}. Qwen2.5-VL further shows strong capabilities on visually rich documents, including pages with layout, tables, charts, and diagrams~\citep{bai2025qwen25vl}.

Historical document studies suggest that this shift is important for archival material. Multimodal LLMs have shown competitive performance on European historical documents~\citep{greif2025mllm_historical,kim2025early,crosilla2025benchmarking}. CHURRO extends this direction with a historical-text VLM trained and evaluated across many historical corpora~\citep{semnani2025churro}. These studies show that VLMs can be useful for historical OCR. They also show that performance depends on deployment choices such as input granularity, prompting, decoding, and evaluation protocol.

Arabic-specific VLM and OCR systems are also emerging. Qalam targets Arabic OCR and handwriting recognition~\citep{bhatia2024qalam}. QARI-OCR adapts multimodal models for Arabic text recognition and includes diacritised text~\citep{qariocr2025}. Baseer focuses on Arabic document-to-Markdown conversion~\citep{baseer2025}. KITAB-Bench provides a multi-domain Arabic OCR and document understanding benchmark~\citep{heakl2025kitab}. SARD provides a synthetic printed Arabic OCR benchmark~\citep{sard2025}.

These systems show rapid progress, but they do not fully answer the manuscript question. Most evaluations focus on modern printed Arabic, controlled handwriting, synthetic Arabic, or general document understanding. Historical Arabic work has mainly produced datasets and supervised HTR systems. The zero-shot behaviour of VLMs on Islamic manuscript images remains underexplored, especially when the same model must be compared against traditional OCR and OCR-conditioned correction across different document settings.

\subsection{OCR Correction and Deployment Choices}
\label{sec_background_correction}

A separate line of work uses language models to correct noisy OCR output. The basic idea is that OCR provides a rough textual prior, and the language model revises it. In historical documents, this can improve recognition when the OCR output still preserves useful anchors~\citep{greif2025mllm_historical}.

Correction is not always beneficial. \citet{kanerva2025ocr} show that LLM-based OCR correction can help in some languages and fail in others. \citet{levchenko2025historical} also reports cases where post-OCR correction degrades historical OCR output. Reference-based correction has been used for historical Vietnamese diacritics~\citep{do2024reference}, but that setting assumes a stronger textual anchor than noisy OCR alone.

These findings motivate a practical question for Arabic manuscripts. A noisy OCR prior may help a VLM by constraining generation and giving it partial text anchors. It may also hurt by pulling the model toward systematic OCR errors. The outcome likely depends on how recoverable the OCR prior is, how compatible it is with the script, how degraded the image is, and which VLM is used as the corrector. This paper studies that question directly through an OCR-conditioned VLM pipeline that receives both the source image and the Tesseract output.

\subsection{Layout, Metadata, and Semantic Understanding}
\label{sec_background_structure}

Text recognition is only one part of manuscript understanding. Digitisation projects also need layout analysis, metadata extraction, entity recognition, and content classification. These tasks support search, cataloguing, linking, and scholarly discovery.

Layout analysis for Arabic-script manuscripts has received less attention than OCR. BADAM provides a public dataset for baseline detection in Arabic-script manuscripts~\citep{badam2019}. Related systems based on document layout models can detect text regions, but most do not model Islamic manuscript structures. Marginal glosses, nested commentary, rubrication, colophons, and paratextual zones remain difficult.

Metadata and entity extraction are also underdeveloped for manuscript images. Arabic NER has been studied on text~\citep{zaghouani2012renar}. Image-based entity extraction has been explored for European historical documents~\citep{greif2025mllm_historical}. To our knowledge, there is no equivalent evaluation for historical Arabic manuscript images.

Text-level Islamic NLP has advanced in parallel. Recent work addresses Qur'anic and hadith QA, hallucination detection, Arabic and Islamic cultural benchmarks, and Arabic-centric generative models~\citep{bhatia-etal-2026-rag,palmx2025,fanar2025}. These systems usually assume clean digital text. They do not solve the image-level bottleneck. The gap between manuscript images and text-level Islamic NLP therefore remains substantial.

\paragraph{Positioning of This Work.}
\label{sec_background_positioning}

Table~\ref{tbl_litmap} summarises the closest prior work. Existing studies cover European historical OCR, Arabic OCR, supervised Arabic HTR, OCR correction, and Arabic document understanding. To our knowledge, prior work has not jointly evaluated zero-shot VLM recognition, OCR-conditioned correction, diacritics sensitivity, and cross-dataset OCR-prior recoverability for historical Arabic manuscript images. This is the gap addressed by our study.

\begin{table*}[t]
\caption{Positioning of this work against related studies. The columns indicate script family, VLM setting, whether OCR correction is tested, whether diacritics are analysed, and whether cross-dataset quality comparison is included. \ding{51}\,=\,yes, \ding{55}\,=\,no, --\,=\,not applicable.}
\label{tbl_litmap}
\setlength{\tabcolsep}{2pt} 
{\scriptsize%
\begin{tabular*}{\textwidth}{@{\extracolsep{\fill}}>{\raggedright\arraybackslash}p{2.3cm}>{\raggedright\arraybackslash}p{2.7cm}cccc@{}}
\toprule
\textbf{Study} & \textbf{Script} & \textbf{VLM setting} & \textbf{Corrector} & \textbf{Diacritics} & \textbf{Cross-dataset} \\
\midrule
\multicolumn{6}{l}{\textit{\textbf{European and multilingual historical OCR}}} \\
\citet{greif2025mllm_historical} & Latin & ZS & \ding{51} & \ding{55} & \ding{55} \\
\citet{kim2025early} & Latin & ZS & \ding{55} & \ding{55} & \ding{55} \\
\citet{crosilla2025benchmarking} & Latin multi & ZS & \ding{55} & \ding{55} & \ding{51} \\
\citet{semnani2025churro} & 46 languages & FT & \ding{55} & \ding{55} & \ding{51} \\
\citet{kanerva2025ocr} & Latin and Finnish & -- & \ding{51} & \ding{55} & \ding{51} \\
\citet{levchenko2025historical} & Latin & ZS & \ding{51} & \ding{55} & \ding{55} \\
\citet{do2024reference} & Vietnamese & -- & \ding{51} & \ding{51} & \ding{55} \\
\midrule
\multicolumn{6}{l}{\textit{\textbf{Arabic OCR and HTR}}} \\
\citet{chan2024hatformer} & Arabic historical & -- & \ding{55} & \ding{55} & \ding{55} \\
\citet{qariocr2025} & Arabic printed & FT & \ding{55} & \ding{51} & \ding{55} \\
\citet{baseer2025} & Arabic modern & FT & \ding{55} & \ding{55} & \ding{55} \\
\citet{heakl2025kitab} & Arabic multi-domain & ZS & \ding{55} & \ding{55} & \ding{51} \\
\midrule
\textbf{This work} & \textbf{Arabic historical and multi-domain} & \textbf{ZS} & \ding{51} & \ding{51} & \ding{51} \\
\bottomrule
\end{tabular*}
}
\end{table*}

\section{Datasets}
\label{sec_datasets}

We evaluate eight Arabic text datasets. They cover historical manuscripts, modern handwriting, multi-domain Arabic OCR, synthetic printed text, aged printed books, and student handwriting. Six datasets are public or previously released external resources. Two datasets are in-house collections obtained directly from their originators. Table~\ref{tbl_datasets} summarises the evaluated datasets and also lists three related resources that define adjacent Arabic document-analysis tasks but are not used in our experiments.

We use the datasets as separate evaluation settings rather than as one pooled benchmark. This is important because each dataset tests a different deployment condition. Muharaf and Madinah test historical manuscript recognition. RASAM~2 tests transfer to Maghribi-script manuscripts. KHATT and Students Essays test modern handwriting under controlled and realistic conditions. KITAB-Bench tests multi-domain Arabic OCR. SARD tests clean printed Arabic. QNL Books tests aged printed Arabic from a realistic digitisation setting.

\begin{table}[htbp]
\centering
\caption{Arabic text recognition datasets. Datasets marked with \textsuperscript{*} are used in our experiments.}
\label{tbl_datasets}
\setlength{\tabcolsep}{2pt} 
{\small%
\begin{tabular}{@{}p{0.16\linewidth}p{0.25\linewidth}p{0.17\linewidth}p{0.25\linewidth}p{0.10\linewidth}@{}}
\toprule
\textbf{Dataset} & \textbf{Content} & \textbf{Scale} & \textbf{Annotation} & \textbf{Access} \\
\midrule
Muharaf\textsuperscript{*} & Historical Arabic manuscripts, 19th--21st c. & ${\sim}$1,600 pages & Line + polygon & Public \\
RASAM~2\textsuperscript{*} & Maghribi manuscripts & ${\sim}$300 pages & Line, PAGE-XML & Public \\
KHATT\textsuperscript{*} & Modern Arabic handwriting & 1,000 writers & Paragraph & Restricted \\
Madinah\textsuperscript{*} & Ancient manuscripts, 8 books & 40 pages & Page & Public \\
KITAB-Bench\textsuperscript{*} & Multi-domain Arabic & 3,760 samples & Mixed, 13 subsets & Public \\
SARD\textsuperscript{*} & Synthetic printed Arabic & 996 samples & Character + word & Public \\
QNL Books\textsuperscript{*} & Aged printed Arabic, in-house & 199 pages / 100 books & Page, line-aligned & In-house \\
Students Essays\textsuperscript{*} & Student handwriting, Grade 4--12, in-house & 200 essays & Page, dual ground truth & In-house \\
MADCAT & Arabic handwriting & ${\sim}$42,000 pages & Page + line & Restricted \\
HistoryAr & Historical Arabic & ${\sim}$100 pages & Line & Public \\
BADAM & Baseline detection & ${\sim}$400 pages & Baseline polygon & Public \\
\bottomrule
\end{tabular}
}
\end{table}

\subsection{External Datasets}
\label{sec_external_datasets}

\emph{Muharaf} is our primary line-level historical manuscript benchmark~\citep{muharaf2024}. We use its standard test split of 1{,}334 line images. The corpus contains more than 1{,}600 Arabic manuscript pages with expert transcriptions and polygon annotations. The script is predominantly Naskh. We use Muharaf for OCR versus VLM comparison, preprocessing analysis, diacritics analysis, and OCR-conditioned correction.

\emph{Madinah} provides 40 page-level images from eight historical Arabic manuscripts~\citep{madinah2024}. We use it to test page-level manuscript recognition and to examine whether VLMs benefit from full-page context. We also use the first and last page of each book as a small candidate set for the metadata extraction pilot, since title pages and colophons are often found near the beginning or end of manuscripts.

\emph{RASAM~2} contains Maghribi Arabic manuscript images from the BULAC Library~\citep{rasam2024}. We use 159 page images with PAGE-XML line-level ground truth. This dataset tests whether Arabic VLM-OCR and OCR-conditioned correction transfer beyond Naskh-style manuscripts. It is therefore important for evaluating script mismatch and OCR-prior recoverability.

\emph{KHATT} is a modern Arabic offline handwriting dataset~\citep{khatt2012}. We use 1{,}038 test paragraphs from controlled writers. It serves as a modern handwriting control condition. Compared with Students Essays, KHATT is more regular because the writing comes from a controlled data collection setup.

\emph{SARD} is a synthetic printed Arabic OCR dataset~\citep{sard2025}. We use 996 samples rendered across six fonts. It represents the clean printed setting. This setting is useful because traditional OCR is expected to be strong, which makes it possible to test whether VLMs and OCR-conditioned correction still add value when the OCR prior is already favourable.

\emph{KITAB-Bench} is a multi-domain Arabic OCR benchmark with 3{,}760 samples across 13 subsets~\citep{heakl2025kitab}. The subsets cover printed text, synthetic text, handwriting, historical documents, and literary Arabic. We use KITAB-Bench in two ways. First, it provides an aggregate test of the OCR-corrector pipeline across diverse Arabic document types. Second, its subsets provide a more fine-grained test of when the OCR prior helps or hurts VLM correction.

\subsection{In-House Collections}
\label{sec_inhouse_datasets}

\emph{QNL Books} is an unpublished in-house collection of aged printed Arabic pages. It contains 199 evaluable page images from 100 Arabic books in the Qatar National Library collection. The pages were originally curated in 2017 for an OCR benchmarking project. The books were printed with older printing technology and show visible aging. Common issues include page yellowing, ink degradation, and mild scanning artefacts.

QNL Books occupies a middle point between clean printed Arabic and historical handwriting. It is harder than SARD because the pages are aged and visually degraded. It is easier than handwritten manuscript data because the text is printed. Each page was transcribed verbatim by a trained linguist with line-level alignment. The transcription preserves diacritics, page numbers, and headers. The original test split contains two pages per book. One page contained only whitespace and was excluded.

\emph{Students Essays} is an unpublished in-house corpus of 200 Arabic essays written by students in Grades 4--12. The pages contain natural handwriting from non-expert writers. They include variation in letter shape, spacing, page tilt, erasures, and mid-line corrections. This makes the corpus more realistic than controlled handwriting benchmarks.

Each Students Essays page has two references. The verbatim reference preserves the spelling and grammar written on the page. The corrected reference normalises those errors. We use the verbatim reference as the primary ground truth because it matches the visual content of the page. We report results against the corrected reference as a supplementary signal. This helps show whether a model preserves the written form or silently corrects it.

Sample-level metadata is available for Students Essays, but it is not analysed in this paper. It extends our evaluation to realistic student handwriting and complements KHATT, which contains more controlled handwriting samples.

The two in-house collections serve two roles. They broaden the evaluation to aged printed Arabic and realistic student handwriting. They also reduce the chance that all results are affected by public benchmark exposure during model pretraining. QNL Books and Students Essays were not publicly released at the time of evaluation.\section{Experimental Design}
\label{sec_experimental_design}

This section describes the evaluation protocol. We first summarise how the experiments map to the research questions. We then describe the evaluated systems, OCR-conditioned correction pipeline, preprocessing variants, prompts, inference settings, and evaluation metrics.

\subsection{Evaluation Overview}
\label{sec_evaluation_overview}

The evaluation is organised around the five research questions. \textit{RQ1} compares traditional OCR and standalone VLMs on line-level and page-level manuscript recognition. \textit{RQ2} tests OCR-conditioned correction across manuscript, print, and handwriting settings. \textit{RQ3} uses Muharaf to study whether common preprocessing operations improve VLM recognition. \textit{RQ4} analyses failure modes that affect Arabic VLM-OCR, including diacritics, hallucination, truncation, and repetition loops. \textit{RQ5} uses a small Madinah metadata pilot to study prompt sensitivity rather than full extraction accuracy. Unless stated otherwise, the main results use fixed prompts and decoding settings for each input type. Reruns on SARD and Students Essays are reported as diagnostic experiments because they isolate token-budget truncation and greedy-decoding repetition failures.

\subsection{Models}
\label{sec_models}

\begin{table}[htbp]
\centering
\caption{Summary of evaluated systems. Kraken~\citep{calfa2024} is part of the open-source HTR landscape but is not included in the head-to-head comparison because its default installation does not ship with an Arabic-capable model.}
\label{tbl_models}
\setlength{\tabcolsep}{2pt} 
{\scriptsize
\begin{tabular*}{\textwidth}{@{\extracolsep{\fill}}>{\raggedright\arraybackslash}p{3.7cm}ll>{\raggedright\arraybackslash}p{2.7cm}@{}}
\toprule
\textbf{Model} & \textbf{Paradigm} & \textbf{Parameters} & \textbf{Architecture} \\
\midrule
Tesseract 5~\citep{smith2007tesseract} & Traditional OCR & --- & Rule-based + LSTM \\
Qwen2.5-VL-7B-Instruct~\citep{bai2025qwen25vl} (Qwen) & General-purpose VLM & 7B & Transformer VL \\
Fanar-2-Oryx-IVU~\citep{fanar2025} (Fanar-2) & Arabic-tuned VLM & 7B & Qwen2.5-VL fine-tuned \\
QARI-OCR~\citep{qariocr2025,qwen3vl2025} & OCR-tuned VLM & 4B & Qwen3-VL-4B + LoRA \\
\bottomrule
\end{tabular*}
}
\end{table}

In Table~\ref{tbl_models}, we summarise the models used in our experiments. All local VLM inference uses bfloat16 precision on NVIDIA GPUs. The comparison is framed as a zero-shot deployment study, not as a supervised HTR benchmark. While supervised Arabic HTR systems may perform better when labelled data from the target collection is available, our focus is on systems that can be deployed without collection-specific training. This setting allows us to compare the practical utility of traditional OCR, standalone VLM recognition, and OCR-conditioned VLM correction across diverse Arabic document types.

\subsection{OCR-Corrector Pipeline}
\label{sec_corrector_design}

\begin{figure}[h]
\centering
\includegraphics[width=0.6\columnwidth]{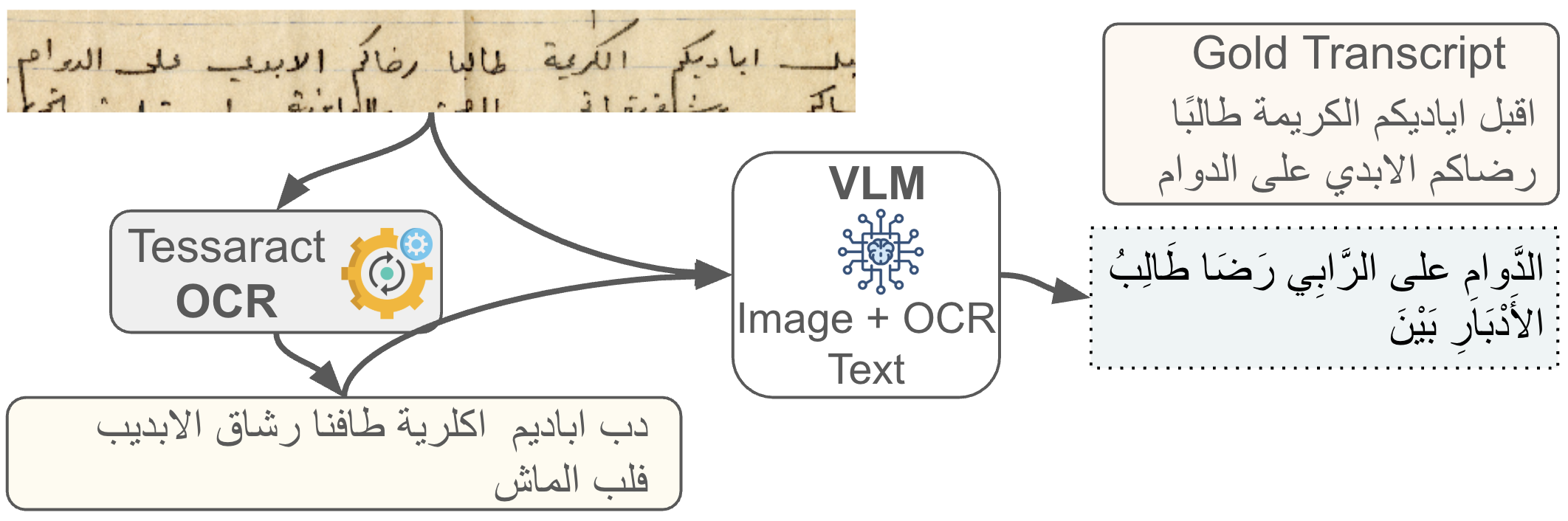}
\caption{Architecture of the hybrid OCR-corrector pipeline. Tesseract provides a noisy text. The VLM receives this prior together with the source image and outputs a transcription. 
}
\label{fig_corrector_arch}
\vspace{-0.3cm}
\end{figure}

We test whether noisy OCR output can act as a useful textual prior for VLM recognition. In Figure~\ref{fig_corrector_arch}, we present the experimental pipeline. It first runs Tesseract on the input image. It then provides the Tesseract output and the source image to a VLM. The VLM is asked to correct the OCR output while checking it against the image. This design follows recent post-OCR correction work on historical documents~\citep{greif2025mllm_historical}, however, it differs in one important way. The corrector receives both the noisy OCR text and the original image.
The OCR-conditioned prompt is shown below.

\begin{quote}
\small\itshape
An OCR system produced the following noisy transcription of this Arabic text image. \texttt{{tesseract\_output}} Please examine the image and produce a corrected transcription. Fix OCR errors while preserving the original text faithfully, including diacritical marks when they are visible. Output only the corrected text.
\end{quote}

This pipeline tests whether OCR can serve as a useful text cue for the VLM. The OCR output may reduce hallucination by giving the model a textual context of the image. The image then helps the model check that text and correct OCR errors. The same prior can also be harmful when Tesseract output is systematically wrong. We therefore evaluate the corrector across different scripts, input granularities, and document settings. Qwen is the primary corrector. Fanar-2 is also evaluated as a second corrector on Muharaf, QNL Books, RASAM~2, and Students Essays. Madinah is excluded from this experiment as it is used for page-level recognition and the metadata analysis.

\subsection{Image Preprocessing}
\label{sec_preprocessing}

To address RQ3, we evaluate whether simple image preprocessing improves VLM recognition. We conduct this experiment on the released Muharaf line images using Qwen, since Muharaf provides expert-transcribed line-level data for historical handwritten Arabic manuscripts.
We evaluate eight input conditions. The \textbf{Raw} condition uses the original image without preprocessing. \textbf{Binarised} applies adaptive binarisation. \textbf{Contrast} applies CLAHE-based contrast enhancement. \textbf{Denoise} uses Gaussian denoising. \textbf{Sharpen} uses unsharp masking. We also test two sequential variants. \textbf{Contrast+Sharpen} applies contrast enhancement followed by sharpening, and \textbf{Denoise+Contrast} applies denoising followed by contrast enhancement. The \textbf{Full} condition applies all preprocessing operations in sequence.
This design tests whether operations that are common in OCR pipelines also help VLMs. It also tests whether preprocessing removes small visual cues that are important for Arabic script, including dots, diacritics, and thin connecting strokes.

\subsection{Prompting Strategy}
\label{sec_prompting}

We keep prompting simple and consistent across models so that differences in output are driven mainly by the model and input condition rather than by prompt variation. For standalone recognition, all VLMs receive the same transcription prompt. The prompt, as presented below, asks for faithful transcription only and suppresses explanations, translation, and commentary.

\begin{quote}
\small\itshape
You are an expert Arabic manuscript transcriber. Read the Arabic text in this image and transcribe it exactly as written. Output only the transcribed Arabic text, nothing else.
\end{quote}

For page-level images, we add only one instruction, namely to transcribe the page line by line. We do not ask the model to normalise, interpret, summarise, or correct the text in the standalone setting. We used English prompts as several prior works report that models are better with English prompts~\citep{kmainasi2024native,bhatia-etal-2026-rag}.
We used greedy decoding for the main recognition experiments to make the outputs reproducible. Diagnostic reruns are reported separately when a different decoding constraint is applied.

\subsection{Experimental Settings}
\label{sec_inference_settings}

All VLM inference was conducted with the \texttt{transformers} library on NVIDIA H100 80~GB GPUs. The default line-level setting uses \texttt{max\_new\_tokens=512}, \texttt{temperature=0}, and \texttt{do\_sample=False}. Page-level inputs often require longer generations. We therefore use \texttt{max\_new\_tokens} values of 2{,}048 or 4{,}096 for Madinah, RASAM~2, QNL Books, Students Essays, and the SARD long-reference rerun.
Students Essays exposes a repetition-loop failure under greedy decoding. For the standalone VLM reruns on the catastrophic subset, we set \texttt{no\_repeat\_ngram\_size=4}. This setting is reported only where it is used and is not silently applied to all results.
We run Tesseract using page segmentation mode 7 for Muharaf line images. We use page segmentation mode 6 or 3 for paragraph-level and page-level inputs. Average inference time is below 0.1 seconds per image for Tesseract, about 2 seconds for standalone VLM inference, and about 3 seconds for the OCR-corrector pipeline.

\subsection{Evaluation Metrics}
\label{sec_metrics}

\paragraph{Text recognition metrics.}
We report Character Error Rate (CER) and Word Error Rate (WER). Both are computed as edit distance normalised by the reference length.

\begin{equation}
\mathrm{CER} =
\frac{d_{\mathrm{char}}(r,h)}
{|r|_{\mathrm{char}}}
\end{equation}

\begin{equation}
\mathrm{WER} =
\frac{d_{\mathrm{word}}(r,h)}
{|r|_{\mathrm{word}}}
\end{equation}

Here $r$ is the reference, $h$ is the hypothesis, $d_{\mathrm{char}}$ is character-level edit distance, and $d_{\mathrm{word}}$ is word-level edit distance. Lower values indicate better recognition. We report scores both with and without Arabic diacritics. The normalised variant strips Arabic diacritical marks from both reference and hypothesis before scoring. The stripped ranges are U+0610--U+061A, U+064B--U+065F, U+0670, and U+06D6--U+06ED. We also unify alef variants and remove tatweel.

\paragraph{Corpus-level aggregation.}
Corpus-level CER and WER are computed by summing edit distances across all samples and dividing by the total reference length. This gives a length-weighted score and prevents short samples from dominating the result. We report bootstrap 95\% confidence intervals with 10{,}000 iterations for the main CER comparisons. We use paired bootstrap tests for key paired comparisons.

\paragraph{Relative improvement.}
For the corrector pipeline, relative improvement is computed against the better standalone method on the same dataset. The standalone methods are Tesseract and the corresponding standalone VLM.

\begin{equation}
\Delta_{\mathrm{rel}} =
\frac{
\mathrm{CER}_{\mathrm{best}} -
\mathrm{CER}_{\mathrm{corr}}
}
{\mathrm{CER}_{\mathrm{best}}}
\times 100
\end{equation}

Here $\mathrm{CER}_{\mathrm{best}}$ is the CER of the better standalone method, and $\mathrm{CER}_{\mathrm{corr}}$ is the CER of the OCR-conditioned corrector. Positive values mean that the corrector improves over the best standalone method. Negative values mean that it underperforms. When we report improvement over Tesseract specifically, we name Tesseract as the baseline.

\paragraph{Error taxonomy.}
We use a lightweight error taxonomy to compare failure modes. The goal is descriptive analysis rather than a complete forensic decomposition of CER. The categories are not mutually exclusive at the character level.

\begin{itemize}[noitemsep,topsep=0em,leftmargin=1.5em,labelsep=.5em]
\item \textbf{Substitution} character-level substitutions reported by \texttt{jiwer.process\_characters} after diacritics stripping, alef unification, and tatweel removal.

\item \textbf{Diacritics} the difference between character-alignment errors computed with and without diacritics stripping.

\item \textbf{Omission} a length-ratio heuristic applied when the prediction is shorter than half the reference.

\item \textbf{Hallucination} a length-ratio heuristic applied when the prediction is longer than 1.5 times the reference.

\end{itemize}

For omission, we attribute the missing characters as $\max(0, \lvert\mathrm{ref}\rvert - \lvert\mathrm{hyp}\rvert)$. For hallucination, we attribute the excess characters as $\max(0, \lvert\mathrm{hyp}\rvert - \lvert\mathrm{ref}\rvert)$. These two heuristics capture severe empty-output and runaway-generation cases. They understate moderate length mismatch, so the resulting percentages should be interpreted as relative failure indicators.

\paragraph{Metadata extraction.}
For metadata extraction, we report per-field non-null rates and inspect schema compliance. Because the metadata pilot lacks expert field-level ground truth, we treat it as a failure-mode analysis rather than as a full precision and recall evaluation.
\section{Results}
\label{sec_results}

In this section, we report the main experimental findings. We first compare traditional OCR and standalone VLMs for manuscript recognition. We then test OCR-conditioned correction and analyse whether the model uses both the OCR prior and the image. We next examine cross-dataset OCR-prior recoverability. The remaining subsections report diagnostic analyses on diacritics, preprocessing, generation budget, repetition loops, metadata extraction, and error types.

\subsection{Standalone OCR and VLM Recognition}
\label{sec_results_standalone}

\begin{table}[ht]
\caption{Text recognition on the Muharaf test set. Corpus-level CER and WER are reported as percentages. Lower is better. Kraken is omitted because its default installation does not ship an Arabic-capable model and produced empty output on all 1{,}334 lines.}
\label{tbl_main_results}
\centering
\begin{tabular}{lcccc}
\toprule
\multirow{2}{*}{\textbf{Model}} & \multicolumn{2}{c}{\textbf{With diacritics}} & \multicolumn{2}{c}{\textbf{Without diacritics}} \\
\cmidrule(lr){2-3} \cmidrule(lr){4-5}
& CER $\downarrow$ & WER $\downarrow$ & CER $\downarrow$ & WER $\downarrow$ \\
\midrule
\multicolumn{5}{l}{\textit{Traditional OCR}} \\
Tesseract 5 & 69.6 & 103.7 & \textbf{69.2} & 103.6 \\
\midrule
\multicolumn{5}{l}{\textit{Vision-Language Models}} \\
Qwen2.5-VL-7B & 86.7 & 108.0 & 70.4 & 104.6 \\
Fanar-2-Oryx-IVU & 87.7 & 109.0 & 70.8 & 105.6 \\
QARI-OCR v0.4.0 & 194.7 & 232.0 & 167.2 & 230.2 \\
\bottomrule
\end{tabular}
\end{table}

\paragraph{Line-level recognition on Muharaf.}
In Table~\ref{tbl_main_results}, we report CER and WER on the Muharaf test set of 1{,}334 line images. We report results with and without diacritics in both the reference and the hypothesis. With diacritics stripped, Tesseract obtains 69.2\% CER, Qwen obtains 70.4\%, and Fanar-2 obtains 70.8\%. The Tesseract--Qwen difference is small. A paired bootstrap test gives a difference of $-1.3$ with a 95\% CI of [$-8.8$, +5.0], which is not statistically significant.

The result shows that current VLMs are close to Tesseract on zero-shot line-level manuscript recognition, although the absolute error rates remain high. This is expected as the evaluation is zero-shot, while supervised systems such as HATFormer~\citep{chan2024hatformer} are trained on target manuscript data. QARI-OCR performs poorly on Muharaf, with 167.2\% normalised CER. This indicates severe domain mismatch between its OCR tuning data and historical manuscript images. In Section~\ref{sec_results_errors}, we report that its errors are dominated by hallucinated text.

\begin{table}[htbp]
\caption{Comparison between Muharaf line images and Madinah page images. CER and WER are reported without diacritics.}
\label{tbl_cross_granularity}
\centering
\begin{tabular}{lccccc}
\toprule
\multirow{2}{*}{\textbf{Model}} & \multicolumn{2}{c}{\textbf{Muharaf line}} & \multicolumn{2}{c}{\textbf{Madinah page}} & \multirow{2}{*}{$\Delta$CER} \\
\cmidrule(lr){2-3} \cmidrule(lr){4-5}
& CER & WER & CER & WER & \\
\midrule
Tesseract 5 & 69.2 & 103.6 & 51.0 & 85.7 & $-18.2$ \\
Qwen2.5-VL-7B & 70.4 & 104.6 & 40.7 & 80.2 & $-29.7$ \\
Fanar-2 & 70.8 & 105.6 & \textbf{39.6} & \textbf{78.7} & $-31.2$ \\
\bottomrule
\end{tabular}
\end{table}

\paragraph{Page-level recognition on Madinah.}
In Table~\ref{tbl_cross_granularity}, we compare line-level Muharaf with page-level Madinah. This comparison does not isolate page context because the datasets also differ in content, image quality, and annotation unit. It is still useful because it shows how model behaviour changes when the input contains a full manuscript page.

On Madinah, Fanar-2 obtains the best CER at 39.6\%. Qwen follows closely at 40.7\%. Both VLMs outperform Tesseract, which obtains 51.0\%. The improvement from Muharaf to Madinah is larger for VLMs than for Tesseract. This suggests that VLMs may benefit from page context, including surrounding text, layout cues, and visual continuity that are lost in line crops. The result should be interpreted cautiously because Madinah contains only 40 pages from 8 books. The Fanar--Qwen gap is 1.1 and is not statistically significant under book-level cluster bootstrap.

\subsection{OCR-Conditioned Correction and Input Ablations}
\label{sec_results_corrector}

We next evaluate whether noisy OCR output can help a VLM. The corrector pipeline provides Tesseract output and the source image to the VLM. The model is asked to produce a corrected transcription.
The best result is Tesseract$\rightarrow$Qwen at 60.6\% CER, as reported in Table~\ref{tbl_corrector}. This is an 8.6 absolute reduction from Tesseract alone and a 12.4\% relative improvement. A paired bootstrap test confirms that the improvement is significant. The 95\% CI is [5.8, 10.9], with $p < 0.001$. Tesseract$\rightarrow$Fanar-2 also improves over the standalone systems and reaches 63.5\% CER. On Muharaf, Qwen is the stronger corrector. This does not hold everywhere, since Fanar-2 is the stronger corrector on QNL Books.
The corpus-level gain hides sample-level variation. The corrector improves CER by more than 1\% on 36.4\% of Muharaf samples and worsens CER by more than 1\% on 44.2\%. The aggregate gain comes from improvements on longer samples, which carry more weight in corpus-level CER. The corrector is therefore useful on average, but it is not risk-free for every line.

\begin{table}[htbp]
\caption{OCR-conditioned correction on Muharaf. Metrics are corpus-level percentages with diacritics normalised.}
\label{tbl_corrector}
\centering
\small
\begin{tabular}{lccc}
\toprule
\textbf{Condition} & \textbf{CER} $\downarrow$ & \textbf{WER} $\downarrow$ & $\Delta$CER \\
\midrule
\multicolumn{4}{l}{\textit{Standalone systems}} \\
Tesseract 5 & 69.2 & 103.6 & --- \\
Qwen2.5-VL-7B (Qwen) & 70.4 & 104.6 & +1.2 \\
Fanar-2-Oryx-IVU (Fanar-2) & 70.8 & 105.6 & +1.6 \\
\midrule
\multicolumn{4}{l}{\textit{OCR-conditioned correctors}} \\
Tesseract $\rightarrow$ Qwen & \textbf{60.6} & \textbf{93.9} & $-8.6$ \\
Tesseract $\rightarrow$ Fanar-2  & 63.5 & 96.3 & $-5.7$ \\
\bottomrule
\end{tabular}
\end{table}

\paragraph{Evidence for OCR-text use.}
The corrector appears to use the OCR prior rather than ignore it. We test this on the 1{,}198 Muharaf lines where Tesseract produced non-empty output. The corrector output is closer to the Tesseract prior than standalone Qwen is in 71.4\% of these samples. The mean normalised distance from Tesseract to the corrector is 133.8\%. The mean distance from Tesseract to standalone Qwen is 178.1\%.

\begin{table}[htbp]
\caption{Ablations on the use of the OCR prior on Muharaf. The ablations use 1{,}198 lines with non-empty Tesseract output. CER is normalised.}
\label{tbl_anchoring}
\centering
\small
\begin{tabular}{@{}p{0.56\linewidth}cp{0.28\linewidth}@{}}
\toprule
\textbf{Condition} & \textbf{CER} & \textbf{Edit-distance signal} \\
\midrule
Tesseract alone & 69.2 & --- \\
Qwen standalone & 70.4 & --- \\
Tesseract $\rightarrow$ Qwen with real prior and image & \textbf{60.6} & baseline \\
\quad Shuffled prior with real image & 81.6 & Distance to wrong prior 64.5\% \\
\quad Real prior with image withheld & 73.9 & Distance to real prior 29.9\% \\
\bottomrule
\end{tabular}
\end{table}

We also run two ablations. The \textit{first} pairs each image with a shuffled Tesseract prior from another sample. The \textit{second} withholds the image and gives the model only the real prior. In Table~\ref{tbl_anchoring}, we report the results.
Both inputs are necessary. Shuffling the prior raises CER from 60.6\% to 81.6\%. The wrong prior actively harms the model. Withholding the image raises CER to 73.9\%. The model then stays much closer to the OCR prior, but produces worse output than Tesseract alone. The corrector therefore depends on both signals. The prior provides an initial textual cue, and the image grounds the correction.

The corrector also reduces the diacritics penalty. Tesseract$\rightarrow$Qwen has an 8.9\% diacritics penalty, compared with 16.3\% for standalone Qwen. Since Tesseract rarely emits diacritics, its output may discourage the VLM from generating incorrect \textit{tashkeel}. This reduces, but does not solve, the diacritics problem.

\subsection{Cross-Dataset OCR-Prior Recoverability}
\label{sec_results_recoverability}

The Muharaf result raises a broader question. Does OCR-conditioned correction help across document settings. Table~\ref{tbl_quality_gradient} compares the best standalone system with the best corrector across seven dataset-level scenarios.
The corrector improves performance on SARD, KITAB-Bench, Muharaf, and QNL Books, however, it gives limited gains or degrades recognition on RASAM~2, KHATT, and Students Essays. This supports an OCR-prior recoverability pattern. OCR helps when its output contains usable text cues that the VLM can verify against the image. It becomes less useful when the OCR text is systematically misleading or mismatched to the script style.

\begin{figure}[htbp]
\centering
\includegraphics[width=0.45\columnwidth]{\detokenize{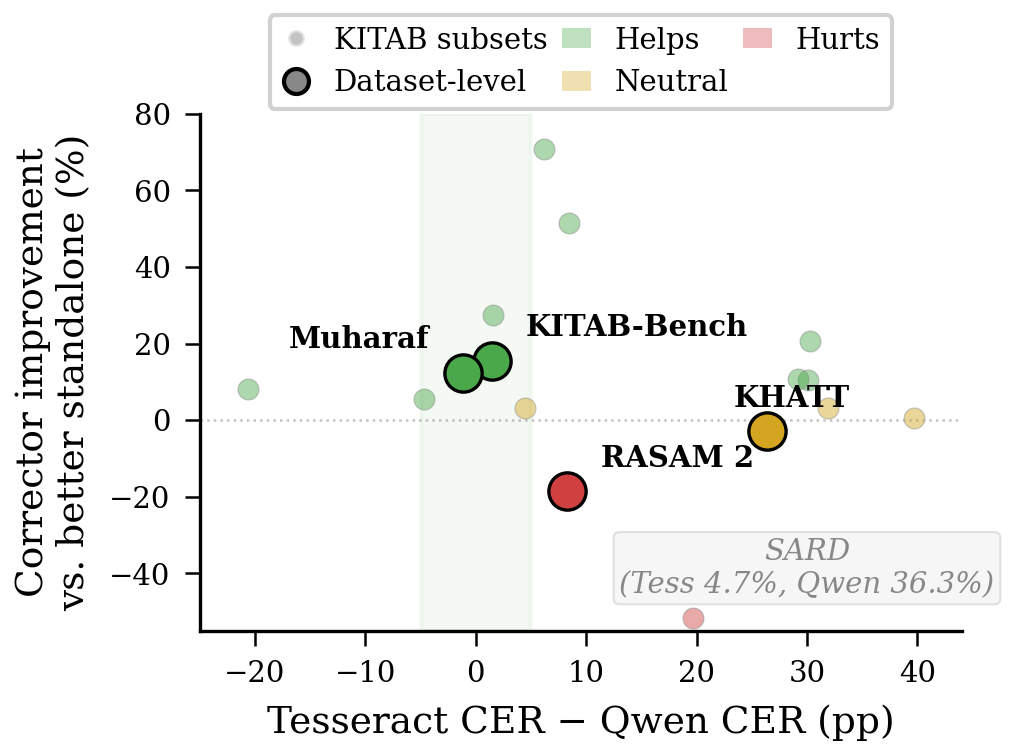}}
\caption{Relative improvement of the best corrector over the best standalone method. The Tesseract--VLM gap is useful, but it is not a complete decision rule. OCR-prior recoverability also depends on script, document setting, and error type.}
\label{fig_quality_gradient}
\end{figure}

The CER gap between Tesseract and the best VLM is useful (Figure~\ref{fig_quality_gradient}), however, it is not enough by itself. QNL Books has a large positive gap of 13.9\%, yet the corrector helps. RASAM~2 has a smaller gap of 8.3\%, yet the corrector hurts. The decision therefore cannot rely only on the gap. Script compatibility, error type, and corrector-model choice also matter. QNL Books shows that aged print can produce noisy yet recoverable OCR. RASAM~2 shows the opposite pattern. Maghribi script appears to produce an OCR prior that misleads the VLM during correction.
The practical implication is cautious, however, it is useful. A small probe sample should compare Tesseract, one or two standalone VLMs, and an OCR-conditioned corrector. The CER gap should be inspected together with length ratio, script type, and qualitative error patterns. If the OCR output preserves usable text cues, the corrector is likely to help. If the OCR output is strongly distorted or script-mismatched, the standalone VLM is safer.

\begin{table*}[t]
\caption{Cross-dataset OCR-prior recoverability analysis. CER values are normalised percentages. The gap is Tesseract CER minus the best standalone VLM CER. Positive relative change means that the corrector improves over the best standalone method.
Students Essays uses per-sample medians. The standalone VLM value uses the mitigated outputs on the catastrophic subset. Corrector medians are unmitigated. All other rows use dataset aggregates. SARD uses the 2{,}048-token rerun.
}
\label{tbl_quality_gradient}
\centering
\footnotesize
\begin{tabular}{llccccc}
\toprule
\textbf{Dataset} & \textbf{Setting} & \textbf{Tess.} & \textbf{Best VLM} & \textbf{Best corr.} & \textbf{$\Delta$Rel.} & \textbf{Gap} \\
\midrule
SARD & Clean printed & 4.7 & Qwen 3.8 & T$\rightarrow$Q \textbf{2.6} & +31.6\% & +0.9 \\
KITAB-Bench & Mixed & 40.2 & Qwen 38.7 & T$\rightarrow$Q \textbf{32.7} & +15.5\% & +1.5 \\
Muharaf & Historical Naskh & 69.2 & Qwen 70.4 & T$\rightarrow$Q \textbf{60.6} & +12.4\% & $-1.2$ \\
RASAM~2 & Historical Maghribi & 72.2 & Qwen 63.9 & T$\rightarrow$Q 75.8 & $-18.6$\% & +8.3 \\
QNL Books & Aged printed & 33.1 & Fanar-2 19.2 & T$\rightarrow$F \textbf{15.5} & +19.3\% & +13.9 \\
Students Essays & Modern student HW & 77.4 & Qwen 49.4 & T$\rightarrow$F 86.7 & $-75.5$\% & +28.0 \\
KHATT & Modern HW & 67.3 & Qwen 40.9 & T$\rightarrow$Q 42.1 & $-2.9$\% & +26.4 \\
\bottomrule
\end{tabular}

\end{table*}

\paragraph{KITAB-Bench.}
\label{sec_results_kitab}

KITAB-Bench evaluates whether the corrector pipeline generalises beyond the main manuscript datasets. It contains 3{,}760 samples across 13 Arabic OCR subsets.
The corrector obtains 32.7\% CER (Table~\ref{tbl_kitab_aggregate}). This is a 15.5\% relative improvement over Qwen and an 18.7\% improvement over Tesseract. When the pathological \texttt{khattparagraph} subset is excluded, the corrector obtains 17.8\% CER. This gives a 24.3\% relative improvement over the better standalone method.

\begin{table}[htbp]
\caption{KITAB-Bench aggregate results. Metrics are dataset-level and diacritics normalised.}
\label{tbl_kitab_aggregate}
\centering
\small
\begin{tabular}{@{}lcc@{}}
\toprule
\textbf{Condition} & \textbf{CER} $\downarrow$ & \textbf{WER} $\downarrow$ \\
\midrule
Tesseract 5 & 40.2 & 77.3 \\
Qwen2.5-VL-7B & 38.7 & 57.5 \\
Tesseract $\rightarrow$ Qwen & \textbf{32.7} & \textbf{51.9} \\
\bottomrule
\end{tabular}
\end{table}

\begin{table}[htbp]
\centering
\small
\caption{KITAB-Bench per-subset CER. Values are normalised percentages. Bold marks the best method per subset.}
\label{tbl_kitab_subsets}
\begin{tabular}{@{}lrrrr@{}}
\toprule
\textbf{Subset} & \textbf{N} & \textbf{Tess.} & \textbf{Qwen} & \textbf{Corrector} \\
\midrule
arabicocr & 50 & \textbf{1.2} & 6.7 & 4.7 \\
hindawi & 200 & 16.5 & 21.2 & \textbf{15.6} \\
patsocr & 500 & 17.5 & 11.3 & \textbf{3.3} \\
isippt & 500 & 18.2 & 38.8 & \textbf{16.7} \\
synthesizear & 500 & 30.4 & 22.0 & \textbf{10.7} \\
onlinekhatt & 200 & 61.7 & 31.6 & \textbf{28.3} \\
historicalbooks & 10 & 62.6 & 30.7 & \textbf{29.7} \\
adab & 200 & 63.0 & 61.4 & \textbf{44.5} \\
khatt & 200 & 63.9 & 34.7 & \textbf{31.0} \\
evarest & 800 & 71.5 & 41.2 & \textbf{32.7} \\
muharaf & 200 & 81.3 & \textbf{61.6} & 93.4 \\
historyar & 200 & 86.7 & 47.0 & \textbf{46.8} \\
khattparagraph & 200 & 412.1 & 407.6 & \textbf{394.3} \\
\bottomrule
\end{tabular}
\end{table}

\begin{figure}[htbp]
\centering
\includegraphics[width=0.8\columnwidth]{\detokenize{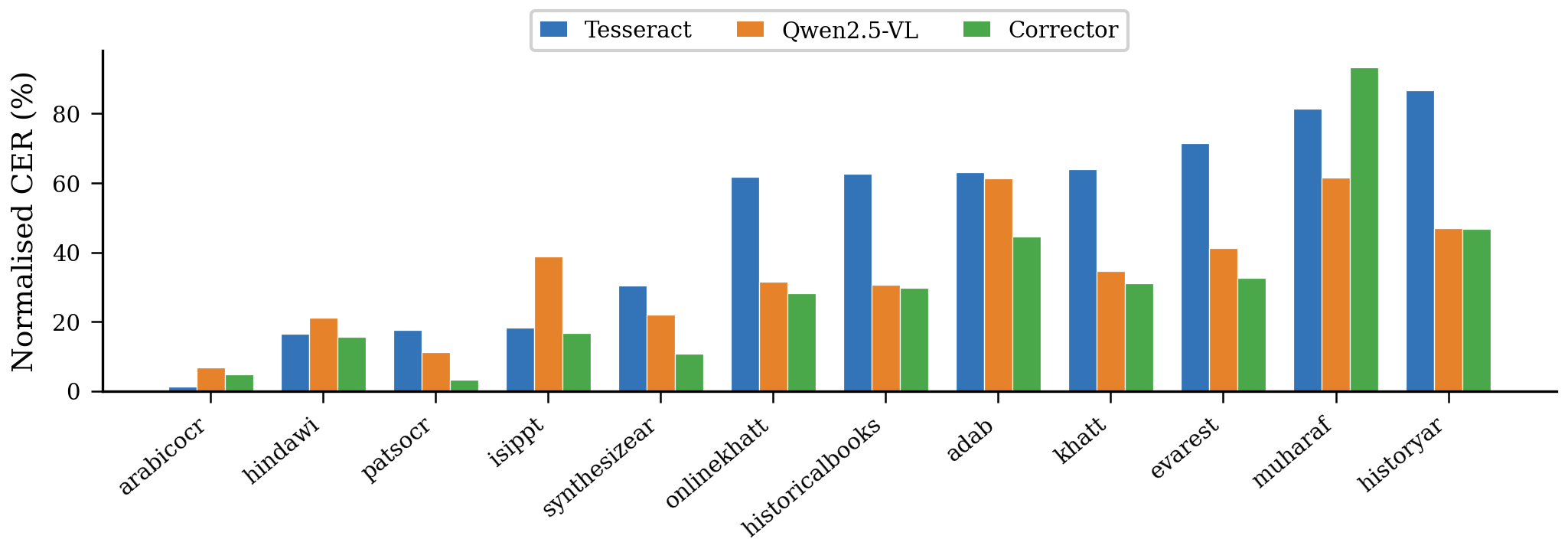}}
\caption{KITAB-Bench per-subset CER. The \texttt{khattparagraph} subset is excluded from the plot because all methods exceed 394\% CER.
}
\label{fig_kitab_subsets}
\end{figure}

The corrector gives the best CER on 11 of 13 subsets (Table~\ref{tbl_kitab_subsets}, visualised in Figure~\ref{fig_kitab_subsets}). One of these wins is not meaningful because all methods exceed 394\% CER on \texttt{khattparagraph}. Excluding that subset, the corrector wins on 10 of 12 non-degenerate subsets. The failure cases are informative. Tesseract wins on \texttt{arabicocr}, a clean printed subset. The standalone VLM wins on the KITAB-Bench \texttt{muharaf} subset. In that subset, Tesseract is much worse than Qwen, and the OCR prior misleads the corrector.

The KITAB-Bench \texttt{muharaf} subset gives lower Qwen CER than our full Muharaf evaluation. We reran our Qwen pipeline on the exact KITAB-Bench 200-sample slice and obtained 61.3\% CER, close to the reported 61.6\%. The difference is therefore due to sample selection rather than pipeline or normalisation differences.

\paragraph{QNL Books.}
\label{sec_results_qnl}

QNL Books tests aged printed Arabic. It sits between clean synthetic print and historical handwriting. The pages are printed, but they show aging, ink degradation, yellowing, and mild scanning artefacts.
Both VLMs outperform Tesseract (Table~\ref{tbl_qnl}). Fanar-2 obtains 19.2\% CER, and Qwen obtains 20.1\%. Tesseract obtains 33.1\%. This shows that aged print is much harder for Tesseract than clean printed SARD. The best result is Tesseract$\rightarrow$Fanar-2 at 15.5\% CER. This is a 19.3\% relative improvement over the best standalone method and a 53.2\% relative improvement over Tesseract. It is the strongest single-dataset gain in the paper.
The QNL result also shows that the corrector model matters. Fanar-2 is a stronger corrector than Qwen on QNL Books, while Qwen is stronger on Muharaf. This may reflect the benefit of Arabic-specialised tuning for aged printed or classical Arabic material. We treat this as a hypothesis because it is based on one dataset.

\begin{table}[htbp]
\caption{QNL Books results. The dataset contains 199 page-level samples from 100 aged printed Arabic books.}
\label{tbl_qnl}
\centering
\small
\begin{tabular}{@{}lcc@{}}
\toprule
\textbf{Condition} & \textbf{CER} $\downarrow$ & \textbf{WER} $\downarrow$ \\
\midrule
Tesseract 5 & 33.1 & 63.9 \\
Qwen2.5-VL-7B & 20.1 & 35.4 \\
Fanar-2-Oryx-IVU & 19.2 & 34.7 \\
Tesseract $\rightarrow$ Qwen & 19.9 & 36.5 \\
Tesseract $\rightarrow$ Fanar-2 & \textbf{15.5} & \textbf{30.0} \\
\bottomrule
\end{tabular}
\end{table}

\subsection{Diacritics Sensitivity}
\label{sec_results_diacritics}

\begin{figure}[htbp]
\centering
\includegraphics[width=0.5\columnwidth]{\detokenize{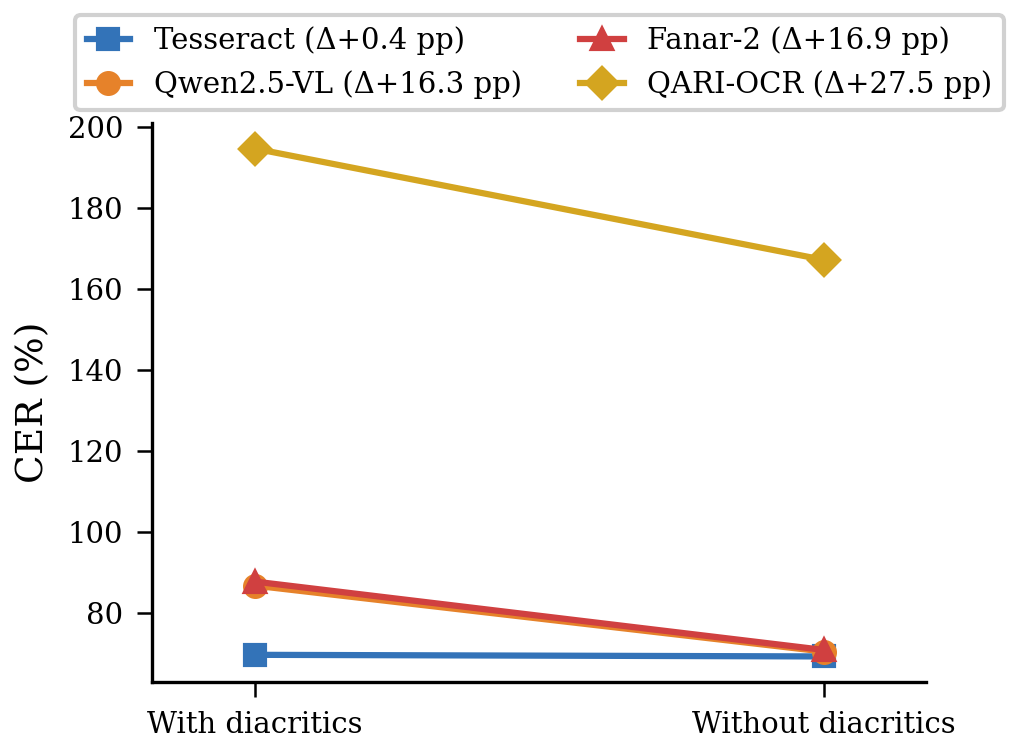}}
\caption{CER with and without \textit{tashkeel}. VLMs show a large diacritics penalty, while Tesseract is nearly flat.}
\label{fig_diacritics}
\end{figure}

Diacritics are a major difference between traditional OCR and VLM recognition. In Table~\ref{tbl_diacritics} and Figure~\ref{fig_diacritics}, we compare CER and WER with and without \textit{tashkeel}.
The main pattern is clear. VLMs are strongly penalised when diacritics are retained. Qwen increases by 16.3\%, and Fanar-2 increases by 16.9\%. Averaged across these two competitive VLMs, the diacritics penalty is 16.6\%. Tesseract behaves differently. Its CER changes by only 0.4\%. This does not mean that Tesseract handles diacritics well. It usually does not produce them. The low penalty reflects omission of \textit{tashkeel}, not faithful preservation.

The same pattern appears beyond Muharaf. In Table~\ref{tbl_diacritics_cross}, we report the Qwen diacritics penalty across five datasets. The penalty is largest on KHATT and Muharaf. It is smaller on SARD, where printed samples contain fewer diacritics. These results motivate reporting diacritics-stripped CER as the main cross-system comparison. Raw CER should still be reported for applications where \textit{tashkeel} fidelity matters.

\begin{table}[htbp]
\caption{Diacritics sensitivity on Muharaf. CER and WER are reported with diacritics retained and with diacritics stripped. The difference is shown in percentage points.}
\label{tbl_diacritics}
\centering
\small
\begin{tabular}{@{}lcccccc@{}}
\toprule
\multirow{2}{*}{\textbf{Model}} & \multicolumn{3}{c}{\textbf{CER}} & \multicolumn{3}{c}{\textbf{WER}} \\
\cmidrule(lr){2-4} \cmidrule(lr){5-7}
& \textbf{+D} & \textbf{$-$D} & \textbf{$\Delta$} & \textbf{+D} & \textbf{$-$D} & \textbf{$\Delta$} \\
\midrule
Tesseract 5 & 69.6 & 69.2 & +0.4 & 103.7 & 103.6 & +0.1 \\
Qwen2.5-VL-7B & 86.7 & 70.4 & +16.3 & 108.0 & 104.6 & +3.4 \\
Fanar-2 & 87.7 & 70.8 & +16.9 & 109.0 & 105.6 & +3.4 \\
QARI-OCR v0.4.0 & 194.7 & 167.2 & +27.5 & 232.0 & 230.2 & +1.8 \\
\bottomrule
\end{tabular}
\end{table}

\begin{table}[htbp]
\caption{Cross-dataset diacritics penalty for Qwen2.5-VL-7B. The values show the CER increase when diacritics are retained. Tesseract is shown where available.}
\label{tbl_diacritics_cross}
\centering
\small
\begin{tabular}{@{}lcc@{}}
\toprule
\textbf{Dataset} & \textbf{Qwen $\Delta$pp} & \textbf{Tess. $\Delta$pp} \\
\midrule
KHATT & +20.2 & --- \\
Muharaf & +16.3 & +0.4 \\
RASAM~2 & +9.8 & +0.1 \\
KITAB-Bench & +6.9 & +1.5 \\
SARD & +2.9 & +2.3 \\
\bottomrule
\end{tabular}
\end{table}

\subsection{Impact of Preprocessing}
\label{sec_results_preprocessing}

\begin{figure}[htbp]
\centering
\includegraphics[width=0.9\columnwidth]{\detokenize{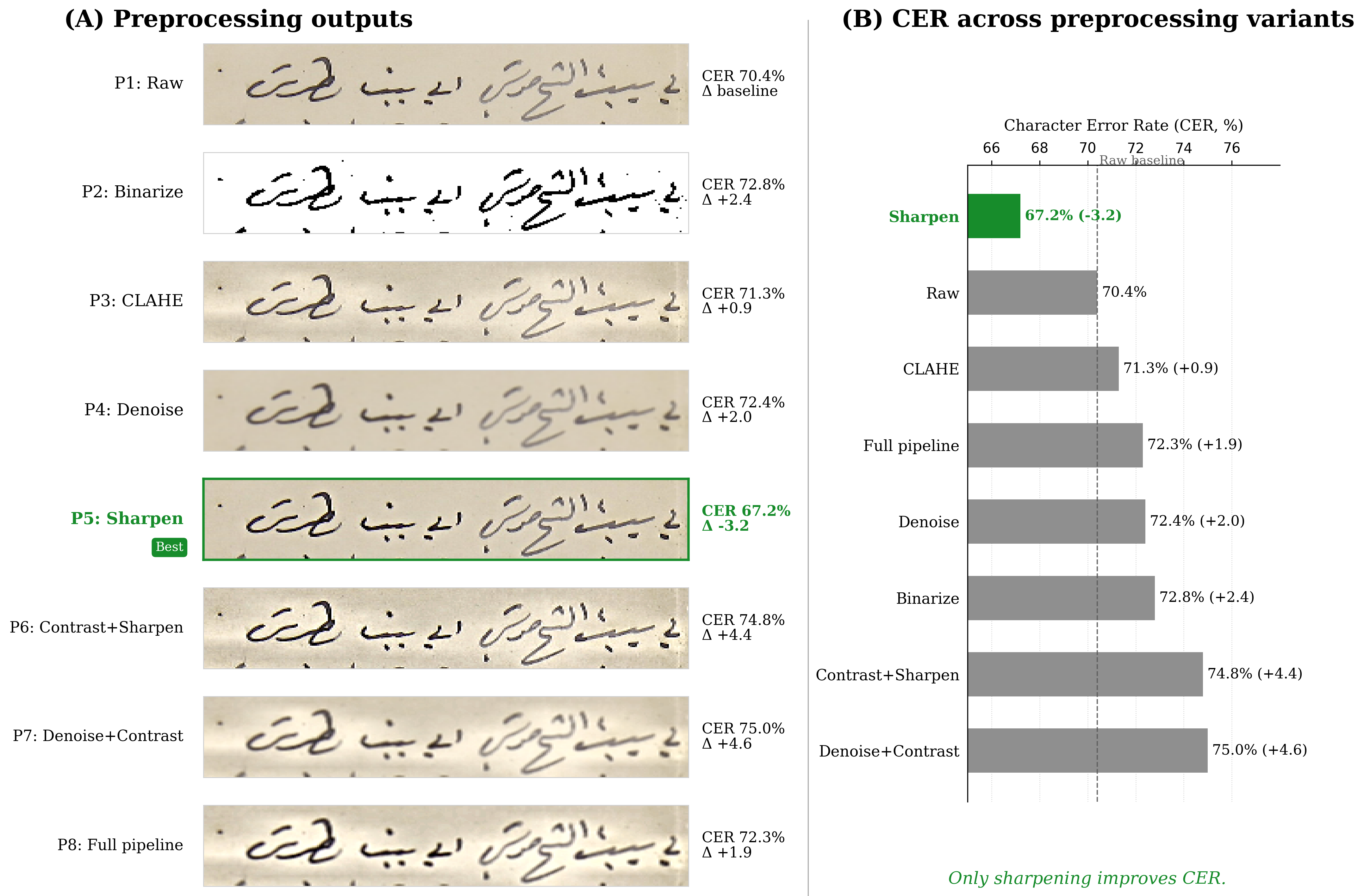}}
\caption{The eight preprocessing variants applied to a Muharaf line crop. Only sharpening improves CER. Other operations either smooth or quantise fine Arabic-script cues.}
\label{fig_preprocessing_pipeline}
\end{figure}

For this experiment we use Qwen on Muharaf line images. As presented in Table \ref{tbl_preprocessing} and in Figure \ref{fig_preprocessing_pipeline}, only sharpening helps. It reduces Qwen CER from 70.4\% to 67.2\%. This makes sharpened Qwen better than standalone Tesseract on Muharaf line images. The effect is VLM-specific. Applying the same sharpening to Tesseract gives 69.5\% CER, which is close to the raw Tesseract result.
Other preprocessing operations degrade performance. Denoising, binarisation, and combined pipelines likely remove or distort fine visual cues. These cues are important in Arabic script, where dots, diacritics, and thin strokes distinguish characters. The result shows that preprocessing is not uniformly helpful for Arabic VLM-OCR. A narrow sharpening operation can help, but more aggressive processing can damage the visual evidence.

\begin{table}[htbp]
\caption{Effect of preprocessing on Qwen2.5-VL-7B. Muharaf results are corpus-level and diacritics normalised.}
\label{tbl_preprocessing}
\centering
\small
\begin{tabular}{@{}lccc@{}}
\toprule
\textbf{Preprocessing} & \textbf{CER} $\downarrow$ & \textbf{WER} $\downarrow$ & \textbf{$\Delta$CER} \\
\midrule
P1 Raw baseline & 70.4 & 104.6 & --- \\
P5 Sharpen & \textbf{67.2} & \textbf{100.6} & $-4.5$\% \\
P3 Contrast CLAHE & 71.3 & 103.3 & +1.3\% \\
P8 Full pipeline & 72.3 & 109.7 & +2.7\% \\
P4 Denoise & 72.4 & 104.0 & +2.8\% \\
P2 Binarised & 72.8 & 106.3 & +3.4\% \\
P6 Contrast+Sharpen & 74.8 & 112.7 & +6.3\% \\
P7 Denoise+Contrast & 75.0 & 112.3 & +6.5\% \\
\bottomrule
\end{tabular}
\end{table}

\subsection{Decoding-Length and Repetition Analysis}
\label{sec_results_generation}

The SARD and Students Essays results show two complementary page-level failure modes. A restrictive maximum output length can truncate long predictions. A larger output limit, however, can give greedy decoding more opportunity to enter repetition loops. We therefore report these cases as targeted analyses of output-length and decoding failures.

\paragraph{SARD.}
\label{sec_results_sard}

SARD evaluates clean printed Arabic. It contains 996 samples rendered across six fonts. The first run used \texttt{max\_new\_tokens}=512. This produced a misleading negative result for VLMs. We therefore report both the original run and a rerun with \texttt{max\_new\_tokens}=2048 (Table~\ref{tbl_sard}).
Under the 512-token setting, Qwen obtains 36.3\% CER and the corrector obtains 36.7\%. Both appear far worse than Tesseract at 4.7\%. The per-font analysis in Table~\ref{tbl_sard_font} shows that this is a truncation artefact. The worst results occur on Arial and Calibri, which have the longest references. Their prediction-to-reference length ratio falls to 0.48.

\begin{table}[htbp]
\caption{SARD aggregate results before and after increasing the VLM generation token size. Tesseract is unaffected by this setting.}
\label{tbl_sard}
\centering
\small
\begin{tabular}{@{}lcccc@{}}
\toprule
\multirow{2}{*}{\textbf{Condition}} & \multicolumn{2}{c}{\textbf{512 tokens}} & \multicolumn{2}{c}{\textbf{2048 tokens}} \\
\cmidrule(lr){2-3} \cmidrule(lr){4-5}
& \textbf{CER} $\downarrow$ & \textbf{WER} $\downarrow$ & \textbf{CER} $\downarrow$ & \textbf{WER} $\downarrow$ \\
\midrule
Tesseract 5 & 4.7 & 24.4 & 4.7 & 24.4 \\
Qwen2.5-VL-7B & 36.3 & 40.8 & 3.8 & 11.2 \\
Tesseract $\rightarrow$ Qwen & 36.7 & 43.7 & \textbf{2.6} & \textbf{12.8} \\
\bottomrule
\end{tabular}
\end{table}

\begin{table}[htbp]
\caption{SARD per-font CER under the original 512-token size. The truncation artefact is concentrated on the long-reference fonts.}
\label{tbl_sard_font}
\centering
\small
\begin{tabular}{@{}lccc@{}}
\toprule
\textbf{Font} & \textbf{Tess.} & \textbf{Qwen 512} & \textbf{Corrector 512} \\
\midrule
Arial & 3.9 & 51.9 & 52.7 \\
Calibri & 4.0 & 52.7 & 53.0 \\
Scheherazade New & 4.4 & 12.5 & 13.7 \\
Sakkal Majalla & 4.5 & 27.1 & 28.4 \\
Traditional Arabic & 4.5 & 29.2 & 28.5 \\
Amiri & 7.7 & 24.0 & 23.2 \\
\bottomrule
\end{tabular}
\end{table}

Increasing the maximum output length to 2{,}048 tokens changes the conclusion. Qwen improves from 36.3\% to 3.8\% CER, while the corrector improves from 36.7\% to 2.6\% CER and slightly outperforms Tesseract. This indicates that the clean printed setting does not reflect a fundamental VLM recognition failure. Instead, page-level VLM-OCR is strongly affected by output-length constraints. Since CER can obscure truncation effects, page- and paragraph-level evaluations should also report prediction-to-reference length ratios.

\paragraph{Students Essays.}
\label{sec_results_students}

Students Essays evaluates realistic student handwriting. It contains 200 Arabic essays from Grades 4--12, with both verbatim and corrected references for each page. We use the verbatim reference as the primary ground truth as it most directly matches the written page.
Median CER favours standalone VLMs over Tesseract. Qwen obtains 55.8\% median CER, compared with 77.4\% for Tesseract. The corrector degrades performance, consistent with the OCR-prior recoverability analysis: Tesseract is much weaker than the VLM on this dataset, so its output provides a poor prior.

\begin{figure}[htbp]
\centering
\includegraphics[width=\columnwidth]{\detokenize{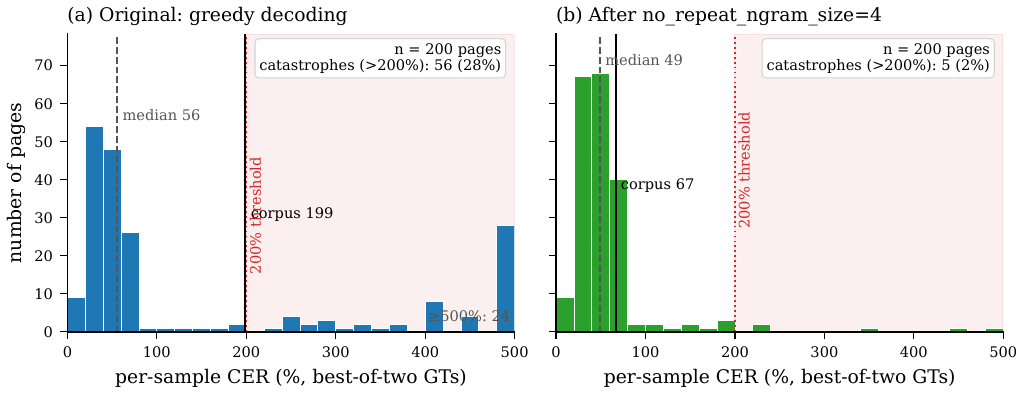}}
\caption{Per-sample CER on Students Essays under Qwen2.5-VL-7B before and after setting \texttt{no\_repeat\_ngram\_size}=4. The mitigation reduces the catastrophic tail and lowers corpus CER from 199.0\% to 66.6\%.}
\label{fig_bimodal}
\end{figure}

However, dataset-level VLM scores are dominated by catastrophic failure cases as reported in Table \ref{tbl_students}. Qwen reaches 199.0\% corpus CER, and Fanar-2 reaches 183.3\%. These large values are caused by repetition loops in a substantial subset of outputs. 28\% of Qwen outputs and 27.5\% of Fanar-2 outputs are excessively long repetitive predictions, which dominate dataset-level CER.
The repetition-loop failure is associated with greedy decoding. Catastrophic outputs have very low trigram diversity, indicating repeated text generation rather than ordinary recognition errors. Rerunning these cases with \texttt{no\_repeat\_ngram\_size}=4 corrects 51 of 56 Qwen failures and 50 of 55 Fanar-2 failures. After replacing the failed outputs with the constrained-decoding outputs, corpus CER drops to 66.6\% for Qwen and 66.9\% for Fanar-2 (Figure~\ref{fig_bimodal}). SARD and Students Essays therefore expose two distinct decoding risks. Page-level Arabic VLM-OCR requires a sufficiently large maximum output length to avoid truncation, while handwriting recognition also requires repetition control to prevent degenerate outputs.

\begin{table}[htbp]
\caption{Students Essays results. CER is normalised. The catastrophe rate is the fraction of samples with per-sample CER above 200\%.}
\label{tbl_students}
\centering
\small
\begin{tabular}{@{}lcccc@{}}
\toprule
\textbf{Condition} & \makecell{\textbf{Median}\\\textbf{CER}} & \makecell{\textbf{Catastrophe}\\\textbf{rate}} & \makecell{\textbf{Corpus}\\\textbf{CER}} & \makecell{\textbf{Corpus CER}\\\textbf{+ norep}} \\
\midrule
Tesseract 5 & 77.4 & 0.0\% & 76.3 & --- \\
Qwen2.5-VL-7B & 55.8 & 28.0\% & 199.0 & \textbf{66.6} \\
Fanar-2-Oryx-IVU & 56.7 & 27.5\% & 183.3 & \textbf{66.9} \\
Tesseract $\rightarrow$ Qwen & 90.9 & 30.5\% & 228.6 & --- \\
Tesseract $\rightarrow$ Fanar-2 & 86.7 & 27.0\% & 208.3 & --- \\
\bottomrule
\end{tabular}
\end{table}

\subsection{Zero-Shot Metadata Extraction}
\label{sec_results_metadata}

We evaluate metadata extraction on 16 candidate pages from Madinah. The pages are the first and last page of each of the eight books. This selection is based on the expectation that title pages and colophons often occur near the beginning or end of a manuscript.

The goal of the evaluation is to see whether prompt design changes VLM behaviour in a structured extraction task. The naive prompt asks the model to extract metadata fields if present. It produces non-null responses for every field on every page (Table~\ref{tbl_metadata}, naive columns), with values that are chunks of transcribed body text rather than actual metadata. This shows a compliance-bias failure mode: the model fills the schema even when the requested information may not be present.
A calibrated prompt reduces this behaviour (Table~\ref{tbl_metadata}, calibrated columns). It forbids extracting prose chunks, requires null when evidence is absent, warns that many pages are body-text pages, and restricts dates and script labels to valid types.

\begin{table}[htbp]
\caption{Zero-shot metadata extraction from 16 Madinah manuscript pages using Qwen, under a naive prompt and a null-enforcing calibrated prompt. Non-null rate measures how often the model attempts an extraction; under the naive prompt this rate is 100\% for every field (compliance bias), while the calibrated prompt returns null for most content fields.}
\label{tbl_metadata}
\centering
\small
\begin{tabular}{@{}lccccc@{}}
\toprule
\multirow{2}{*}{\textbf{Field}} & \multicolumn{2}{c}{\textbf{Naive prompt}} & \multicolumn{2}{c}{\textbf{Calibrated prompt}} & \multirow{2}{*}{\textbf{Category}} \\
\cmidrule(lr){2-3} \cmidrule(lr){4-5}
 & Non-null & Rate & Non-null & Rate & \\
\midrule
\multicolumn{6}{l}{\textit{Visual classification}} \\
Language     & 16/16 & 100.0 & 16/16 & 100.0 & Classification \\
Script type  & 16/16 & 100.0 & 16/16 & 100.0 & Classification \\
\midrule
\multicolumn{6}{l}{\textit{Content extraction}} \\
Subject     & 16/16 & 100.0 & 3/16 & 18.8 & Content \\
Author      & 16/16 & 100.0 & 2/16 & 12.5 & Content \\
Title       & 16/16 & 100.0 & 1/16 & 6.2  & Content \\
Place       & 16/16 & 100.0 & 1/16 & 6.2  & Content \\
Date Hijri  & 16/16 & 100.0 & 0/16 & 0.0  & Content \\
Date CE     & 16/16 & 100.0 & 0/16 & 0.0  & Content \\
\bottomrule
\end{tabular}
\end{table}

The calibrated prompt returns language and script type for all pages. It suppresses most content-field extraction. Content fields have non-null rates between 0\% and 18.8\%. These residual outputs may include correct extractions or remaining compliance bias. We cannot separate these cases without expert annotation. The robust finding is therefore not metadata extraction accuracy. It is prompt sensitivity. A permissive prompt causes systematic field filling, while a null-enforcing prompt substantially reduces this behaviour.

\subsection{Failure Taxonomy}
\label{sec_results_errors}

\begin{figure}[htbp]
\centering
\includegraphics[width=0.6\columnwidth]{\detokenize{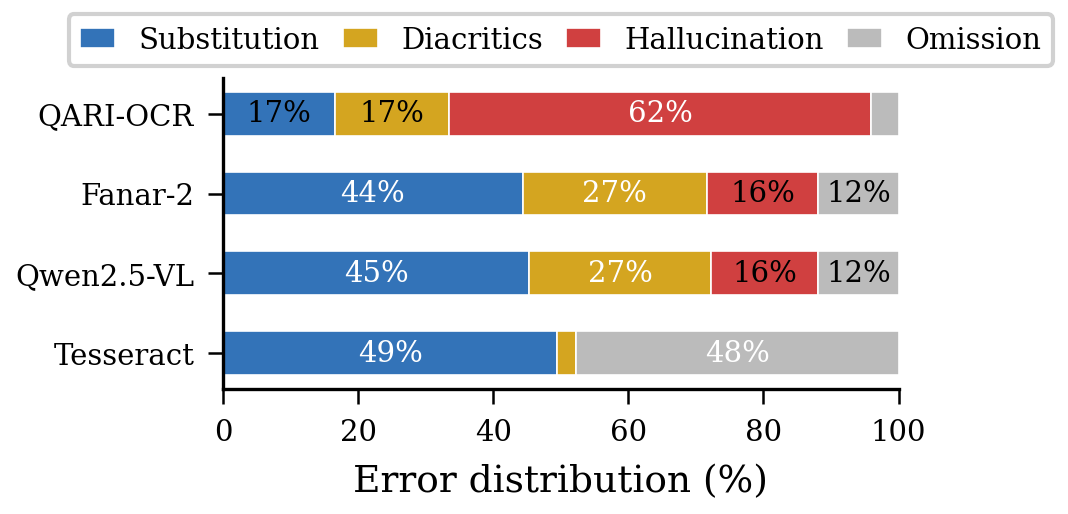}}
\caption{Error type distribution on Muharaf. Tesseract errors are dominated by substitution and omission. VLM errors shift toward diacritics and hallucination.}
\label{fig_error_taxonomy}
\end{figure}

Finally, we compare how the systems fail on Muharaf. In Table~\ref{tbl_errors} and Figure \ref{fig_error_taxonomy}, we report relative error attributions. The categories are heuristic and should not be read as a complete decomposition of CER.
\textit{(i)} Tesseract errors are dominated by substitution and omission. Together they account for 97.2\% of its attributed errors. This reflects the classic behaviour of an OCR pipeline on out-of-domain cursive script.
\textit{(ii)} VLM errors have a different profile. Qwen and Fanar-2 have similar substitution rates to Tesseract, however, much higher diacritics and hallucination rates. Diacritics account for 27.0\% of Qwen errors and 27.3\% of Fanar-2 errors. Hallucination accounts for 15.9\% and 16.5\%, respectively.
\textit{(iii)} VLMs omit less text than Tesseract. Their omission rate is 11.9\%, compared with 47.7\% for Tesseract. This suggests a coverage--precision trade-off. VLMs are more willing to produce text, however, this increases hallucination risk.
\textit{(iv)} QARI-OCR shows the strongest hallucination pattern. Hallucination accounts for 62.4\% of its attributed errors. This is consistent with the high CER reported in Table~\ref{tbl_main_results}. It suggests that Arabic OCR tuning alone does not guarantee robustness on historical manuscript images.

\begin{table}[htbp]
\caption{Heuristic error type attribution on Muharaf. Each column is normalised so that the four heuristic counts sum to 100\%.}
\label{tbl_errors}
\centering
\small
\begin{tabular}{@{}lcccc@{}}
\toprule
\textbf{Error type} & \textbf{Tess.} & \textbf{Qwen} & \textbf{Fanar-2} & \textbf{QARI} \\
\midrule
Substitution & 49.4 & 45.3 & 44.3 & 16.6 \\
Omission & 47.7 & 11.9 & 11.9 & 4.1 \\
Diacritics & 2.7 & 27.0 & 27.3 & 16.9 \\
Hallucination & 0.1 & 15.9 & 16.5 & 62.4 \\
\bottomrule
\end{tabular}
\end{table}

\section{Discussion}
\label{sec_discussion}

The results answer the research questions at three levels. They show when zero-shot VLMs are useful, when OCR-conditioned correction helps or hurts, and which evaluation choices matter for Arabic manuscript digitisation.

\subsection{RQ1. How do VLMs compare with traditional OCR and HTR?}
\label{sec_discussion_rq1}

Zero-shot VLMs are not yet replacements for supervised HTR, yet they are no longer only weak baselines. On line-level Muharaf, Qwen approaches Tesseract after diacritics normalisation. On page-level Madinah, VLMs outperform Tesseract, suggesting that page context can improve recognition through neighbouring text, layout cues, and visual continuity.

This has practical implications for deployment. Traditional OCR relies mainly on local character and word evidence, whereas VLMs can exploit broader visual context. This is useful when pages show repeated letter forms, stable writing style, or layout regularities. Manuscript projects should therefore consider full-page or region-level inference when context is informative. Line crops remain important for controlled evaluation, although they may understate the practical value of VLMs.

Supervised HTR remains the stronger option when labelled training data from the target collection is available. Zero-shot VLM-OCR serves a different setting, where collection-specific transcription is costly or unavailable. This setting is common in the long tail of Islamic manuscript collections.

\subsection{RQ2. When does OCR-conditioned VLM correction help?}
\label{sec_discussion_rq2}

OCR-conditioned correction is useful when the OCR prior is recoverable. The prior need not be highly accurate, however, it must preserve enough readable structure for the VLM to refine using the image. This explains the gains on Muharaf, KITAB-Bench, SARD after increasing the maximum output length, and QNL Books. In these settings, Tesseract output is noisy, however, it still provides useful text cues, while the image allows the VLM to verify and correct them. The failures follow the same pattern: on RASAM~2, KHATT, and Students Essays, the OCR prior is too misleading and degrades correction. In these cases, the standalone VLM is the safer option.

The CER gap between Tesseract and the best VLM is informative but not sufficient. QNL Books has a larger gap than RASAM~2, yet the corrector helps on QNL Books and degrades performance on RASAM~2. This suggests that the type of OCR error matters: aged print can yield noisy but recoverable OCR, whereas Maghribi script can produce OCR text that is less compatible with the image evidence. The input ablations further support this interpretation: \textit{(i)} shuffling the OCR prior degrades the corrector, and \textit{(ii)} removing the image also degrades the corrector. Thus, the corrector is not simple text-only post-editing. It uses both textual and visual evidence. The OCR prior provides partial text cues, and the image validates and corrects them.

The practical implication is adaptive model selection. A digitisation project should first probe the target collection with Tesseract, one or two standalone VLMs, and an OCR-conditioned corrector. The final system should then be selected based on CER, prediction-to-reference length ratio, script type, and qualitative OCR error patterns.

\subsection{RQ3. Does preprocessing improve VLM recognition?}
\label{sec_discussion_rq3}

The preprocessing results show that generic image enhancement does not uniformly improve VLM recognition. On Muharaf, sharpening improves Qwen2.5-VL-7B, whereas most other operations degrade performance. This finding is important because OCR pipelines often assume that binarisation, denoising, and contrast enhancement are broadly beneficial. Arabic manuscript recognition depends on small visual cues, including dots, diacritics, and thin connecting strokes. Sharpening may make these cues more visible, while denoising or binarisation may remove or distort them. Thus, preprocessing that makes an image appear cleaner may still reduce the information needed by a VLM.

The practical recommendation is targeted preprocessing. Sharpening should be tested on a small probe set, while more aggressive operations should not be assumed beneficial. This is especially important for Arabic manuscripts, where small marks can carry high linguistic value.

\subsection{RQ4. What failure modes shape Arabic VLM-OCR?}
\label{sec_discussion_rq4}

Traditional OCR and VLMs fail in different ways. Tesseract mainly produces substitutions and omissions, whereas VLMs omit less text yet introduce additional risks, including diacritics sensitivity, hallucination, and decoding failures. The diacritics result is especially important for Arabic manuscripts. VLMs often attempt \textit{tashkeel}, although many marks are misplaced. Tesseract avoids much of this penalty because it rarely emits diacritics. This creates an evaluation issue. A model that omits diacritics can appear stronger under raw CER, even when it fails to preserve part of the manuscript text. We therefore report both raw and diacritics-stripped CER. Diacritics-stripped CER better reflects base-text recognition, while raw CER remains necessary for applications where \textit{tashkeel} fidelity matters, such as Quranic and scholarly manuscripts.

The generation results show two complementary page-level risks. SARD shows that a restrictive maximum output length can cause artificial truncation. Students Essays shows that larger output limits can expose repetition loops under greedy decoding. Page-level Arabic VLM-OCR therefore requires both sufficient maximum output length and repetition control. More broadly, VLM-OCR evaluation should not rely only on CER or WER. Evaluations should therefore report prediction-to-reference length ratios, repetition rates, and checks for hallucinated text, since CER and WER alone may miss truncation, repeated text, or invented content.

\subsection{RQ5. What does the metadata pilot show?}
\label{sec_discussion_rq5}

The metadata pilot should be read as a prompt-sensitivity and failure-mode study, not as an accuracy evaluation, because the selected pages lack expert field-level ground truth. The clearest finding is compliance bias. With a permissive prompt, the model filled missing metadata fields with body text. A calibrated prompt reduced this behaviour by making null outputs explicit and expected. This indicates that structured metadata extraction from manuscript images requires strong prompt constraints and explicit support for abstention.

The pilot also suggests a staged workflow. VLMs may be most useful as an early routing component, identifying language, script type, and pages likely to contain titles, colophons, or other cataloguing cues. Content metadata should then be extracted only from pages likely to contain relevant evidence, with filtering or human review before catalogue or knowledge-graph ingestion. This is safer than asking a VLM to extract all fields from arbitrary manuscript pages and is consistent with the OCR results. A reliable manuscript pipeline should route pages using page type, script, OCR-prior recoverability, and output-quality checks.

\subsection{Cross-Cutting Deployment Implications}
\label{sec_discussion_guidance}

Across the research questions, Arabic manuscript recognition is best viewed as adaptive system selection rather than single-model selection. A practical pipeline should begin with a small annotated probe from the target collection, comparing Tesseract, at least one standalone VLM, and an OCR-conditioned corrector. The probe should report CER, prediction-to-reference length ratio, hallucinated text, repetition, script type, and qualitative OCR-prior errors. These signals indicate whether the OCR prior provides useful text cues or misleading input. When the OCR prior remains readable and recoverable, the corrector is preferred because the prior constrains generation and the image supports visual correction. When the OCR prior is strongly distorted, script-mismatched, or dominated by systematic errors, the standalone VLM is safer. This explains why correction helps on aged print, mixed-domain Arabic, clean printed Arabic after increasing the maximum output length, and parts of historical Naskh, but degrades performance on Maghribi manuscripts and realistic student handwriting.

The same adaptive view applies to the surrounding design choices. Full-page or region-level inputs should be used when page context supports recognition, while line crops remain useful for controlled evaluation. Preprocessing should be treated as a targeted intervention, since sharpening may improve fine Arabic-script cues whereas denoising and binarisation may remove them. Evaluation should report diacritics-stripped CER for base-text comparison and raw CER when \textit{tashkeel} fidelity matters. Page-level VLM-OCR also requires sufficient maximum output length and repetition control, since truncation and repetition loops are both realistic failure modes. For metadata extraction, prompts should explicitly allow empty outputs, and extracted fields should be filtered or reviewed before catalogue or knowledge-graph ingestion. Overall, these findings support an OCR-VLM workflow that routes pages by script, page type, OCR-prior recoverability, length diagnostics, and repetition indicators rather than relying on a universal recogniser.

\subsection{Limitations and Future Work}
\label{sec_limitations}

This study is designed as a zero-shot deployment analysis rather than an exhaustive benchmark of all OCR, HTR, and VLM systems. We focus on reproducible local inference and do not evaluate frontier API models such as GPT-4o, Claude, or Gemini. These models may perform better on page-level recognition and structured extraction, however, comparing them raises additional issues of cost, reproducibility, versioning, and long-term accessibility. The OCR-prior recoverability analysis is also conducted mainly at the dataset level. Future work should develop page- or region-level routers that choose between Tesseract, standalone VLMs, and OCR-conditioned correctors using signals such as OCR length ratio, script type, image quality, repetition indicators, and OCR error patterns.

The metadata and error analyses should be interpreted as diagnostic studies rather than complete evaluations. The metadata pilot shows prompt sensitivity and compliance-bias hallucination, but does not measure extraction accuracy because expert field-level annotations are unavailable. Similarly, the error taxonomy is heuristic and does not provide a full alignment-based decomposition of every recognition error. Future benchmarks should include expert metadata labels, manual error audits, layout annotations, unpublished collections, and contamination checks where possible. They should also cover a wider range of scripts and page types, including Naskh, Maghribi, Thuluth, Nasta'liq, marginalia, commentary, title pages, and colophons.
\section{Conclusion}
\label{sec_conclusion}

This paper studied when VLMs should be used for Arabic manuscript recognition and when traditional OCR should guide them. Across eight Arabic text datasets and more than 7,700 samples, we show that no single system is uniformly best. Traditional OCR, standalone VLMs, and OCR-conditioned VLM correction each perform best in different settings. The central finding is an OCR-prior recoverability pattern: correction improves recognition when OCR output remains readable enough to guide the VLM, but degrades performance when the prior is systematically misleading. Controlled ablations show that both the OCR prior and the source image are needed, while cross-domain results show that this pattern holds across historical manuscripts, aged print, clean print, and handwriting. The study also identifies deployment-critical failure modes, including diacritics sensitivity, selective effects of preprocessing, output-length truncation, repetition loops, and metadata hallucination. Overall, the findings provide a practical decision framework for Islamic manuscript collections where supervised HTR training data is costly or unavailable.


\FloatBarrier
\bibliographystyle{plainnat}
\bibliography{references}

\begin{thebibliography}{40}
\providecommand{\natexlab}[1]{#1}
\providecommand{\url}[1]{\texttt{#1}}
\expandafter\ifx\csname urlstyle\endcsname\relax
  \providecommand{\doi}[1]{doi: #1}\else
  \providecommand{\doi}{doi: \begingroup \urlstyle{rm}\Url}\fi

\bibitem[Abdallah et~al.(2024)Abdallah, Eberharter, Pfister, and
  Jatowt]{abdallah2024survey}
Abdelrahman Abdallah, Daniel Eberharter, Zoe Pfister, and Adam Jatowt.
\newblock A survey of recent approaches to form understanding in scanned
  documents.
\newblock \emph{Artificial Intelligence Review}, 57:\penalty0 342, 2024.
\newblock \doi{10.1007/s10462-024-11000-0}.

\bibitem[Alwajih et~al.(2025)Alwajih, El~Mekki, Mubarak, Hawasly, Mohamed, and
  Abdul-Mageed]{palmx2025}
Fakhraddin Alwajih, Abdellah El~Mekki, Hamdy Mubarak, Majd Hawasly, Abubakr
  Mohamed, and Muhammad Abdul-Mageed.
\newblock {PalmX} 2025: The first shared task on benchmarking {LLMs} on
  {Arabic} and {Islamic} culture.
\newblock \emph{arXiv preprint arXiv:2509.02550}, 2025.

\bibitem[Bai et~al.(2025{\natexlab{a}})Bai, Chen, Liu, Wang, Ge, Song, Dang,
  Wang, Wang, Tang, et~al.]{bai2025qwen25vl}
Shuai Bai, Keqin Chen, Xuejing Liu, Jialin Wang, Wenbin Ge, Sibo Song, Kai
  Dang, Peng Wang, Shijie Wang, Jun Tang, et~al.
\newblock {Qwen2.5-VL Technical Report}.
\newblock \emph{arXiv preprint arXiv:2502.13923}, 2025{\natexlab{a}}.

\bibitem[Bai et~al.(2025{\natexlab{b}})]{qwen3vl2025}
Shuai Bai et~al.
\newblock {Qwen3-VL} technical report.
\newblock \emph{arXiv preprint arXiv:2511.21631}, 2025{\natexlab{b}}.

\bibitem[Bhatia et~al.(2024)Bhatia, Nagoudi, Alwajih, and
  Abdul-Mageed]{bhatia2024qalam}
Gagan Bhatia, El~Moatez~Billah Nagoudi, Fakhraddin Alwajih, and Muhammad
  Abdul-Mageed.
\newblock Qalam: A multimodal {LLM} for {A}rabic optical character and
  handwriting recognition.
\newblock In \emph{Proceedings of the Second Arabic Natural Language Processing
  Conference}, pages 210--224, Bangkok, Thailand, 2024. Association for
  Computational Linguistics.
\newblock \doi{10.18653/v1/2024.arabicnlp-1.19}.

\bibitem[Bhatia et~al.(2026{\natexlab{a}})Bhatia, Mubarak, Hawasly, Jarrar,
  Mikros, Zaraket, Alhirthani, Al-Khatib, Cochrane, Darwish, Yahiaoui, and
  Alam]{Bhatia_2026}
Gagan Bhatia, Hamdy Mubarak, Majd Hawasly, Mustafa Jarrar, George Mikros, Fadi
  Zaraket, Mahmoud Alhirthani, Mutaz Al-Khatib, Logan Cochrane, Kareem Darwish,
  Rashid Yahiaoui, and Firoj Alam.
\newblock Advances in ai systems on islamic knowledge capabilities: A critical
  survey.
\newblock \emph{TechRxiv (preprint)}, February 2026{\natexlab{a}}.
\newblock \doi{10.36227/techrxiv.177155997.77147487/v1}.
\newblock URL \url{http://dx.doi.org/10.36227/techrxiv.177155997.77147487/v1}.

\bibitem[Bhatia et~al.(2026{\natexlab{b}})Bhatia, Mubarak, Jarrar, Mikros,
  Zaraket, Alhirthani, al~Khatib, Cochrane, Darwish, Yahiaoui, and
  Alam]{bhatia-etal-2026-rag}
Gagan Bhatia, Hamdy Mubarak, Mustafa Jarrar, George Mikros, Fadi Zaraket,
  Mahmoud Alhirthani, Mutaz al~Khatib, Logan Cochrane, Kareem~Mohamed Darwish,
  Rashid Yahiaoui, and Firoj Alam.
\newblock From {RAG} to agentic {RAG} for faithful islamic question answering.
\newblock In Maria Liakata, Viviane~P. Moreira, Jiajun Zhang, and David
  Jurgens, editors, \emph{Findings of the {A}ssociation for {C}omputational
  {L}inguistics: {ACL} 2026}, pages 26469--26488, San Diego, California, United
  States, July 2026{\natexlab{b}}. Association for Computational Linguistics.
\newblock ISBN 979-8-89176-395-1.
\newblock \doi{10.18653/v1/2026.findings-acl.1317}.
\newblock URL \url{https://aclanthology.org/2026.findings-acl.1317/}.

\bibitem[Borchmann et~al.(2021)Borchmann, Pietruszka, Stanislawek, Jurkiewicz,
  Turski, Szyndler, and Grali{\'n}ski]{borchmann2021measuring}
{\L}ukasz Borchmann, Micha{\l} Pietruszka, Tomasz Stanislawek, Dawid
  Jurkiewicz, Micha{\l} Turski, Karolina Szyndler, and Filip Grali{\'n}ski.
\newblock Measuring the state of document understanding.
\newblock In \emph{NeurIPS 2021 Datasets and Benchmarks}, 2021.

\bibitem[{Calfa}(2024)]{calfa2024}
{Calfa}.
\newblock {Arabic OCR} and {HTR} benchmark, 2024.
\newblock URL \url{https://calfa.fr/ocr-arabic-benchmark/}.
\newblock Accessed: 2026.

\bibitem[Chan et~al.(2024)Chan, Mijar, Saeed, Wong, and
  Khater]{chan2024hatformer}
Adrian Chan, Anupam Mijar, Mehreen Saeed, Chau-Wai Wong, and Akram Khater.
\newblock {HATformer}: Historic handwritten arabic text recognition with
  transformers.
\newblock \emph{arXiv preprint arXiv:2410.02179}, 2024.

\bibitem[Crosilla et~al.(2025)Crosilla, Klic, and
  Colavizza]{crosilla2025benchmarking}
Giorgia Crosilla, Lukas Klic, and Giovanni Colavizza.
\newblock Benchmarking large language models for handwritten text recognition.
\newblock \emph{arXiv preprint arXiv:2503.15195}, 2025.

\bibitem[Do et~al.(2024)Do, Tran, Vo, and Kim]{do2024reference}
Thao Do, Dinh~Phu Tran, An~Vo, and Daeyoung Kim.
\newblock Reference-based post-{OCR} processing with {LLM} for precise
  diacritic text in historical document recognition.
\newblock \emph{arXiv preprint arXiv:2410.13305}, 2024.

\bibitem[Faizullah et~al.(2023)Faizullah, Ayub, Hussain, and
  Khan]{faizullah2023survey}
Safiullah Faizullah, Muhammad~Sohaib Ayub, Sajid Hussain, and Muhammad~Asad
  Khan.
\newblock A survey of {OCR} in {Arabic} language: Applications, techniques, and
  challenges.
\newblock \emph{Applied Sciences}, 13\penalty0 (7):\penalty0 4584, 2023.
\newblock \doi{10.3390/app13074584}.

\bibitem[{FANAR Team} et~al.(2026){FANAR Team}, Abbas, Ahmad, Ahmad, Al-Homaid,
  Al-Nuaimi, Altinisik, Asgari, Chawla, Chowdhury, Dalvi, Darwish, Durrani,
  Elfeky, Elmagarmid, Eltabakh, et~al.]{fanar2025}
{FANAR Team}, Ummar Abbas, Mohammad~Shahmeer Ahmad, Minhaj Ahmad, Abdulaziz
  Al-Homaid, Anas Al-Nuaimi, Enes Altinisik, Ehsaneddin Asgari, Sanjay Chawla,
  Shammur Chowdhury, Fahim Dalvi, Kareem Darwish, Nadir Durrani, Mohamed
  Elfeky, Ahmed Elmagarmid, Mohamed Eltabakh, et~al.
\newblock Fanar 2.0: {Arabic} generative {AI} stack.
\newblock \emph{arXiv preprint arXiv:2603.16397}, 2026.

\bibitem[Gacek(2009)]{gacek2009arabic}
Adam Gacek.
\newblock \emph{Arabic manuscripts: a vademecum for readers}, volume~98.
\newblock Brill, 2009.

\bibitem[Graves and Schmidhuber(2008)]{graves2008offline}
Alex Graves and J\"{u}rgen Schmidhuber.
\newblock Offline handwriting recognition with multidimensional recurrent
  neural networks.
\newblock In \emph{Advances in Neural Information Processing Systems},
  volume~21, 2008.

\bibitem[Greif et~al.(2025)Greif, Griesshaber, and
  Greif]{greif2025mllm_historical}
Gavin Greif, Niclas Griesshaber, and Robin Greif.
\newblock Multimodal {LLMs} for {OCR}, {OCR} post-correction, and named entity
  recognition in historical documents.
\newblock \emph{arXiv preprint arXiv:2504.00414}, 2025.

\bibitem[Gruber(2010)]{gruber2010islamic}
Christiane~J Gruber.
\newblock \emph{The Islamic manuscript tradition: ten centuries of book arts in
  Indiana University collections}.
\newblock Indiana University Press, 2010.

\bibitem[Gupta et~al.(2007)Gupta, Jacobson, and
  Garcia]{gupta2007ocr_binarization}
Maya~R. Gupta, Nathaniel~P. Jacobson, and Eric~K. Garcia.
\newblock {OCR} binarization and image pre-processing for searching historical
  documents.
\newblock \emph{Pattern Recognition}, 40\penalty0 (2):\penalty0 389--397, 2007.
\newblock \doi{10.1016/j.patcog.2006.04.043}.

\bibitem[Heakl et~al.(2025)Heakl, Sohail, Ranjan, Elbadry, Ahmad, El-Geish,
  Maher, Shen, Khan, and Khan]{heakl2025kitab}
Ahmed Heakl, Muhammad~Abdullah Sohail, Mukul Ranjan, Rania Elbadry,
  Ghazi~Shazan Ahmad, Mohamed El-Geish, Omar Maher, Zhiqiang Shen,
  Fahad~Shahbaz Khan, and Salman Khan.
\newblock {KITAB}-bench: A comprehensive multi-domain benchmark for {A}rabic
  {OCR} and document understanding.
\newblock In \emph{Findings of the Association for Computational Linguistics:
  ACL 2025}, pages 22006--22024, Vienna, Austria, 2025. Association for
  Computational Linguistics.
\newblock \doi{10.18653/v1/2025.findings-acl.1135}.

\bibitem[Hennara et~al.(2025)Hennara, Hreden, Hamed, Bastati, Aldallal, Chrouf,
  and AlModhayan]{baseer2025}
Khalil Hennara, Muhammad Hreden, Mohamed~Motasim Hamed, Ahmad Bastati, Zeina
  Aldallal, Sara Chrouf, and Safwan AlModhayan.
\newblock Baseer: A vision-language model for {Arabic} document-to-markdown
  {OCR}.
\newblock \emph{arXiv preprint arXiv:2509.18174}, 2025.

\bibitem[Kanerva et~al.(2025)Kanerva, Ledins, K{\"a}pyaho, and
  Ginter]{kanerva2025ocr}
Jenna Kanerva, Cassandra Ledins, Siiri K{\"a}pyaho, and Filip Ginter.
\newblock {OCR} error post-correction with {LLM}s in historical documents: No
  free lunches.
\newblock In \emph{Proceedings of the Third Workshop on Resources and
  Representations for Under-Resourced Languages and Domains
  (RESOURCEFUL-2025)}, pages 38--47, Tallinn, Estonia, 2025. University of
  Tartu Library, Estonia.

\bibitem[Kasem et~al.(2023)Kasem, Mahmoud, and Kang]{kasem2023arabic}
Mahmoud~SalahEldin Kasem, Mohamed Mahmoud, and Hyun-Soo Kang.
\newblock Advancements and challenges in {Arabic} optical character
  recognition: A comprehensive survey.
\newblock \emph{arXiv preprint arXiv:2312.11812}, 2023.

\bibitem[Kiessling et~al.(2019)Kiessling, {Stökl Ben Ezra}, and
  Miller]{badam2019}
Benjamin Kiessling, Daniel {Stökl Ben Ezra}, and Matthew~Thomas Miller.
\newblock {BADAM}: A public dataset for baseline detection in {Arabic}-script
  manuscripts.
\newblock In \emph{Proceedings of the 5th International Workshop on Historical
  Document Imaging and Processing}, pages 13--18, 2019.
\newblock \doi{10.1145/3352631.3352648}.

\bibitem[Kim et~al.(2022)Kim, Hong, Yim, Nam, Park, Yim, Hwang, Yun, Han, and
  Park]{kim2022donut}
Geewook Kim, Teakgyu Hong, Moonbin Yim, JeongYeon Nam, Jinyoung Park, Jinyeong
  Yim, Wonseok Hwang, Sangdoo Yun, Dongyoon Han, and Seunghyun Park.
\newblock {OCR-Free Document Understanding Transformer}.
\newblock In \emph{Computer Vision -- ECCV 2022}, pages 498--517. Springer,
  2022.
\newblock \doi{10.1007/978-3-031-19815-1_29}.

\bibitem[Kim et~al.(2025)Kim, Baudru, Ryckbosch, Bersini, and
  Ginis]{kim2025early}
Seorin Kim, Julien Baudru, Wouter Ryckbosch, Hugues Bersini, and Vincent Ginis.
\newblock Early evidence of how {LLMs} outperform traditional systems on
  {OCR/HTR} tasks for historical records.
\newblock \emph{arXiv preprint arXiv:2501.11623}, 2025.

\bibitem[Kmainasi et~al.(2024)Kmainasi, Khan, Shahroor, Bendou, Hasanain, and
  Alam]{kmainasi2024native}
Mohamed~Bayan Kmainasi, Rakif Khan, Ali~Ezzat Shahroor, Boushra Bendou, Maram
  Hasanain, and Firoj Alam.
\newblock Native vs non-native language prompting: A comparative analysis.
\newblock In \emph{International Conference on Web Information Systems
  Engineering}, pages 406--420. Springer, 2024.

\bibitem[Levchenko(2025)]{levchenko2025historical}
Maria Levchenko.
\newblock Evaluating {LLMs} for historical document {OCR}: A methodological
  framework.
\newblock \emph{arXiv preprint arXiv:2510.06743}, 2025.

\bibitem[Mahmoud et~al.(2014)Mahmoud, Ahmad, Al-Khatib, Alshayeb, Parvez,
  M\"{a}rgner, and Fink]{khatt2012}
Sabri~A. Mahmoud, Irfan Ahmad, Wasfi~G. Al-Khatib, Mohammad Alshayeb,
  Mohammad~Tanvir Parvez, Volker M\"{a}rgner, and Gernot~A. Fink.
\newblock {KHATT}: An open {Arabic} offline handwritten text database.
\newblock \emph{Pattern Recognition}, 47\penalty0 (3):\penalty0 1096--1112,
  2014.
\newblock \doi{10.1016/j.patcog.2013.08.009}.

\bibitem[Nacar et~al.(2025)Nacar, Al-Habashi, Sibaee, Ammar, and
  Boulila]{sard2025}
Omer Nacar, Yasser Al-Habashi, Serry Sibaee, Adel Ammar, and Wadii Boulila.
\newblock {SARD}: A large-scale synthetic {Arabic OCR} dataset for book-style
  text recognition.
\newblock \emph{arXiv preprint arXiv:2505.24600}, 2025.

\bibitem[Najam and Faizullah(2024)]{madinah2024}
Rayyan Najam and Safiullah Faizullah.
\newblock A scarce dataset for ancient {Arabic} handwritten text recognition.
\newblock \emph{Data in Brief}, 55:\penalty0 110729, 2024.

\bibitem[{Qatar National Library}(2026)]{qnl2026heritage}
{Qatar National Library}.
\newblock Heritage library, 2026.
\newblock URL \url{https://www.qnl.qa/en/explore/heritage-library}.
\newblock Accessed: 2026-02-11.

\bibitem[Quiring-Zoche(2013)]{quiring2013colophon}
Rosemarie Quiring-Zoche.
\newblock The colophon in arabic manuscripts. a phenomenon without a name.
\newblock \emph{Journal of Islamic Manuscripts}, 4\penalty0 (1):\penalty0
  49--81, 2013.

\bibitem[Saeed et~al.(2024)Saeed, Chan, Mijar, Moukarzel, Habchi, Younes,
  Elias, Wong, and Khater]{muharaf2024}
Mehreen Saeed, Adrian Chan, Anupam Mijar, Joseph Moukarzel, Georges Habchi,
  Carlos Younes, Amin Elias, Chau-Wai Wong, and Akram Khater.
\newblock Muharaf: Manuscripts of handwritten {Arabic} dataset for cursive text
  recognition.
\newblock \emph{Advances in Neural Information Processing Systems},
  37:\penalty0 58525--58538, 2024.

\bibitem[Semnani et~al.(2025)Semnani, Zhang, He, Tekgurler, and
  Lam]{semnani2025churro}
Sina Semnani, Han Zhang, Xinyan He, Merve Tekgurler, and Monica Lam.
\newblock {CHURRO}: Making history readable with an open-weight large
  vision-language model for high-accuracy, low-cost historical text
  recognition.
\newblock In Christos Christodoulopoulos, Tanmoy Chakraborty, Carolyn Rose, and
  Violet Peng, editors, \emph{Proceedings of the 2025 Conference on Empirical
  Methods in Natural Language Processing}, pages 34777--34824, Suzhou, China,
  November 2025. Association for Computational Linguistics.
\newblock ISBN 979-8-89176-332-6.
\newblock \doi{10.18653/v1/2025.emnlp-main.1763}.
\newblock URL \url{https://aclanthology.org/2025.emnlp-main.1763/}.

\bibitem[Smith(2007)]{smith2007tesseract}
Ray Smith.
\newblock An overview of the {Tesseract OCR} engine.
\newblock In \emph{Ninth International Conference on Document Analysis and
  Recognition (ICDAR)}, volume~2, pages 629--633. IEEE, 2007.
\newblock \doi{10.1109/ICDAR.2007.4376991}.

\bibitem[Vidal-Gor\`{e}ne et~al.(2024)Vidal-Gor\`{e}ne, Salah, Lucas,
  Decours-Perez, and Perrier]{rasam2024}
Chahan Vidal-Gor\`{e}ne, Cl\'{e}ment Salah, No\"{e}mie Lucas, Ali\'{e}nor
  Decours-Perez, and Antoine Perrier.
\newblock Enhancing {Arabic Maghribi} handwritten text recognition with {RASAM}
  2: A comprehensive dataset and benchmarking.
\newblock In \emph{Computational Humanities Research (CHR 2024)}, volume 3834
  of \emph{CEUR Workshop Proceedings}, pages 200--216, 2024.
\newblock URL \url{https://ceur-ws.org/Vol-3834/paper35.pdf}.

\bibitem[Wasfy et~al.(2025)Wasfy, Nacar, Elkhateb, Reda, Elshehy, Ammar, and
  Boulila]{qariocr2025}
Ahmed Wasfy, Omer Nacar, Abdelakreem Elkhateb, Mahmoud Reda, Omar Elshehy, Adel
  Ammar, and Wadii Boulila.
\newblock {QARI-OCR}: High-fidelity {Arabic} text recognition through
  multimodal large language model adaptation.
\newblock \emph{arXiv preprint arXiv:2506.02295}, 2025.

\bibitem[Wei et~al.(2024)Wei, Liu, Chen, Wang, Kong, Xu, Ge, Zhao, Sun, Peng,
  Han, and Zhang]{wei2024gotocr}
Haoran Wei, Chenglong Liu, Jinyue Chen, Jia Wang, Lingyu Kong, Yanming Xu,
  Zheng Ge, Liang Zhao, Jianjian Sun, Yuang Peng, Chunrui Han, and Xiangyu
  Zhang.
\newblock {General OCR Theory: Towards OCR-2.0 via a Unified End-to-end Model}.
\newblock \emph{arXiv preprint arXiv:2409.01704}, 2024.

\bibitem[Zaghouani(2012)]{zaghouani2012renar}
Wajdi Zaghouani.
\newblock {RENAR}: A rule-based {Arabic} named entity recognition system.
\newblock \emph{ACM Transactions on Asian Language Information Processing},
  11\penalty0 (1):\penalty0 2:1--2:13, 2012.
\newblock \doi{10.1145/2090176.2090178}.

\end{thebibliography}

\end{document}